\documentclass[conference]{IEEEtran}
\usepackage{times}

\usepackage[numbers]{natbib}
\usepackage{multicol}
\usepackage[bookmarks=true]{hyperref}

\usepackage{xcolor}
\usepackage{graphicx}

\usepackage{multirow}
\usepackage[linesnumbered,ruled,vlined]{algorithm2e}
\usepackage{nicefrac}
\usepackage{wrapfig}

\usepackage{makecell}
\usepackage{pifont}
\usepackage{amsmath,amssymb}
\SetKw{Initialize}{Initialize}
\usepackage{booktabs}

\usepackage[T1]{fontenc}
\usepackage[utf8]{inputenc}

\newcommand{\BibTeX}{\rm B\kern-.05em{\sc i\kern-.025em b}\kern-.08em\TeX}

\newcommand{\minor}{\textcolor{black}}

\newcommand{\new}{\textcolor{black}}

\newcommand{\rev}{\textcolor{black}}

\begin{document}

\title{Interactive Distillation for Cooperative Multi-Agent Reinforcement Learning}

\author{Author Names Omitted for Anonymous Review. Paper-ID [add your ID here]}



%
\author{\authorblockN{Minwoo Cho,
Batuhan Altundas, and Matthew Gombolay}
\authorblockA{Institute for Robotics and Intelligent Machines, Georgia Institute of Technology}
\authorblockA{\{mcho318, batuhan\}@gatech.edu,  matthew.gombolay@cc.gatech.edu}}

\maketitle

\begin{abstract}

\rev{Knowledge distillation (KD) has the potential to accelerate MARL by employing a centralized teacher for decentralized students but faces key bottlenecks. Specifically, there are (1) challenges in synthesizing high-performing teaching policies in complex domains, (2) difficulties when teachers must reason in out-of-distribution (OOD) states, and (3) mismatches between the decentralized students’ and the centralized teacher’s observation spaces. To address these limitations, we propose HINT (Hierarchical INteractive Teacher-based transfer), a novel KD framework for MARL in a centralized training, decentralized execution setup. By leveraging hierarchical RL, HINT provides a scalable, high-performing teacher. Our key innovation, pseudo off-policy RL, enables the teacher policy to be updated using both teacher and student experience, thereby improving OOD adaptation. HINT also applies performance-based filtering to retain only outcome-relevant guidance, reducing observation mismatches. We evaluate HINT on challenging cooperative domains (e.g., FireCommander for resource allocation, MARINE for tactical combat). Across these benchmarks, HINT outperforms baselines, achieving improvements of 60\% to 165\% in success rate.}

\end{abstract}


\IEEEpeerreviewmaketitle

\section{Introduction}
Multi-agent reinforcement learning (MARL) has been widely studied across various cooperative tasks, including search and rescue operations~\cite{sar1,sar2}, replenishment at sea~\cite{statisticalRASP, brown2017scheduling}, and warehouse management~\cite{war1, war2}, where agents must coordinate their actions to achieve common goals. As task complexity and team size increase, there is growing demand for methods that enable \rev{effective} coordination without sacrificing individual agent autonomy. Although there have been many advancements, \rev{deploying MARL} remains non-trivial due to challenges including partial observability, non-stationary dynamics, and effective credit assignment.

To address these challenges, centralized training with decentralized execution (CTDE) has become a dominant paradigm in MARL. CTDE utilizes global state information or joint observations during the training phase to stabilize learning, while preserving scalability by restricting agents to local observations during execution. Building on this framework, numerous CTDE-based algorithms have shown success in cooperative benchmarks~\cite{QMIX,qtran, MAPPO, HATRPO_HAPPO}. \rev{Building further}, communication-based approaches~\cite{commnet, IC3Net, hetnet2} have been developed to enhance coordination by enabling agents to share information, infer shared context, and adapt to dynamic conditions. \rev{While this extension greatly improves cooperation, it also complicates training: agents must simultaneously optimize both decision-making and communication, which increases sensitivity to noisy signals and policy gradient variance~\cite{difficult_communication}.} To overcome these limitations, we propose a novel framework that integrates hierarchical supervision with interactive knowledge \minor{distillation} to enable \rev{robust} MARL with minimal coordination overhead.



\begin{figure}[t]
    \centering
    \includegraphics[width=0.46\textwidth]{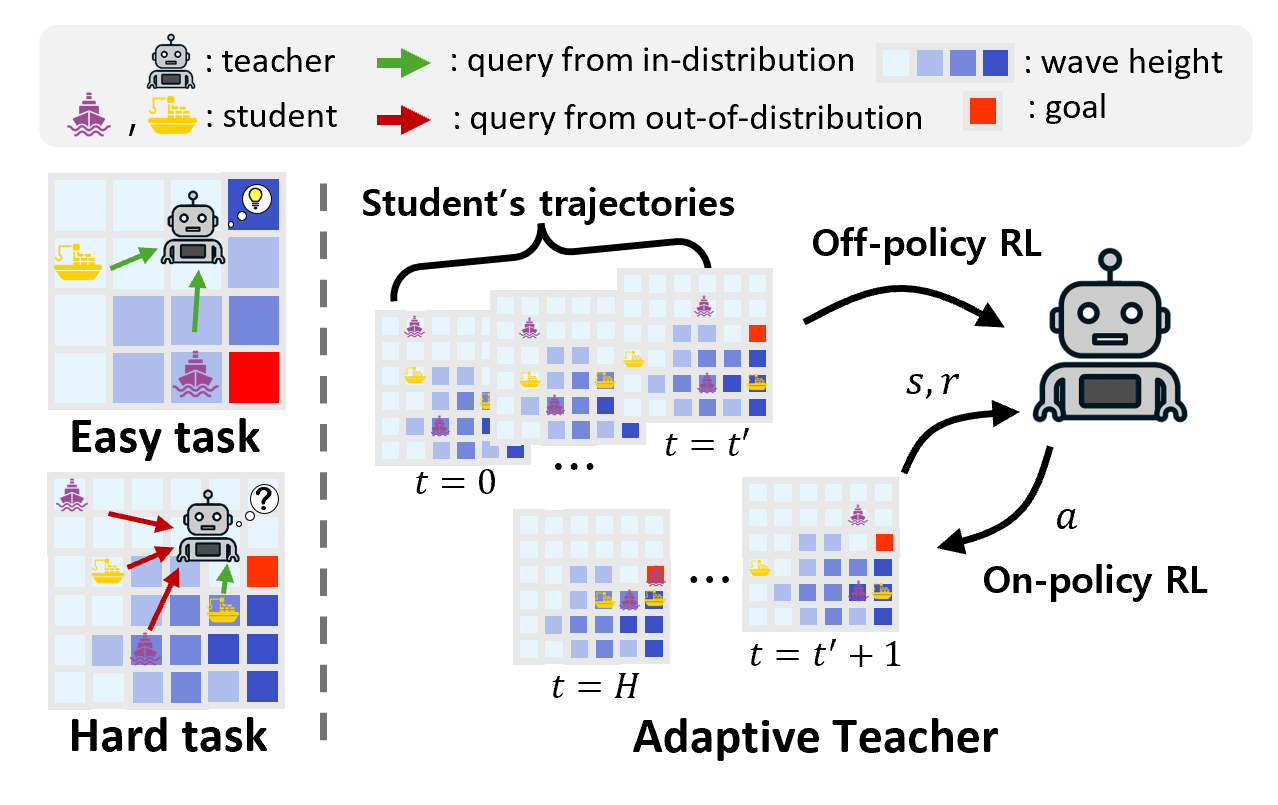}
    \vspace{-14pt}
    \caption{\textbf{Bridging the distribution gap between teacher and student.} As task complexity increases, teachers trained offline provide unreliable guidance on student trajectories that diverge from the training distribution; we close this gap via adaptive refinement. \rev{(Example environment: MARINE)}}
    \label{fig:illustrative_example}
    \vspace{-10pt}
\end{figure}

\begin{figure*}[t]
\centering
\includegraphics[width=0.98\textwidth]
{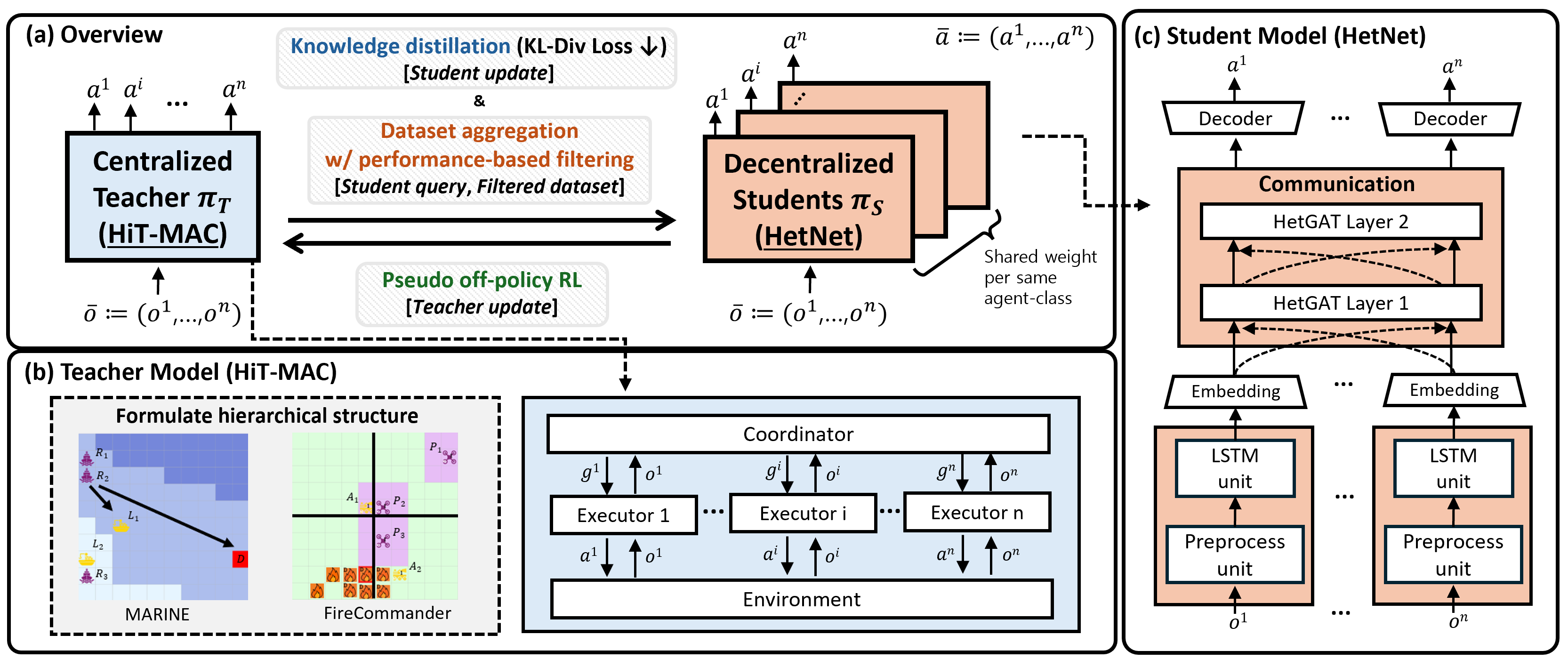}
\vspace{-10pt}
\caption{
\textbf{Overview of \rev{HINT}.}
(a) A centralized teacher guides decentralized students via three mechanisms: \minor{knowledge distillation for student updates, pseudo off-policy RL for teacher refinement, and dataset aggregation with performance-based filtering to support student queries and build a high-quality dataset.}
(b) The teacher operates hierarchically, with a high-level coordinator assigning subgoals to low-level executors based on a task-specific hierarchical structure. This structure enables temporal abstraction by decoupling strategic and tactical decisions. (c) Each student includes a preprocessing unit, an LSTM encoder for temporal abstraction, and a decoder for action selection, while HetGAT layers enable inter-agent communication.}
\label{fig:overview}
\vspace{-10pt}
\end{figure*}

Recently, knowledge distillation (KD)~\cite{kd}, which allows a simpler student model to learn by mimicking the outputs of a larger teacher model, has emerged as a promising alternative to alleviate unstable training signals and the credit assignment problem in MARL~\cite{ctds, ptde, igm-da, cesma}. In this paradigm, a centralized teacher with access to global states supervises student policies by minimizing the divergence between their respective policy distributions. However, distillation approaches face \rev{key bottlenecks}. As task complexity increases, \rev{synthesizing} a reliable, high-performing teacher becomes increasingly \rev{challenging}. While optimal solvers exist for structured tasks like multi-agent pathfinding~\cite{primal, graph_mapf}, most scenarios rely on MARL-based teacher training, which remains brittle and sample-inefficient in \rev{complex}, dynamic domains. Moreover, in such settings, even strong teachers often encounter \rev{out-of-distribution (OOD)} student states, resulting in inconsistent or poor-quality demonstrations (see Fig.~\ref{fig:illustrative_example}, Sec.~\ref{sec:suboptimality}). Finally, asymmetries between teacher and student observations degrade supervision quality—teachers must adapt guidance dynamically to align with student context.

To address these limitations, we propose \rev{HINT (Hierarchical INteractive Teacher-based transfer)} -- a novel \rev{KD} framework \rev{for effective} cooperative \rev{MARL} (Fig.~\ref{fig:overview}). We first pre-train a centralized teacher \rev{with} hierarchical reinforcement learning (RL), decomposing decisions across two levels to \rev{enhance scalability and performance}. Then, decentralized student policies are trained via \rev{KD}, augmented by online expert queries to closely align training data with the student’s test-time distribution. \rev{A key feature of HINT is \textit{pseudo off-policy RL}, in which the teacher is updated using both its own and student trajectories.} This improves awareness of student behavior and enables \rev{richer guidance in OOD states}. \rev{Performance-based filter} further improves dataset quality by retaining only outcome-relevant demonstrations, \rev{reducing observation mismatches}. We test \rev{HINT} in two challenging environments (e.g., FireCommander (FC)~\cite{HetNet1} for resource allocation, MARINE~\cite{ras} for tactical combat), where stochasticity and time-varying dynamics pose additional training difficulties. Experiments show that \rev{HINT} consistently outperforms baselines, validating \rev{its robustness in complex domains}. \textbf{\minor{Our} contributions are as follows:}

\begin{itemize}
  \setlength{\topsep}{3pt}
  \setlength{\itemsep}{0.5pt}
  \setlength{\parsep}{0.1pt}

  \item \rev{We design the teacher with hierarchical RL to manage coordination overheads. This structure improves sample efficiency and teacher effectiveness, making the distillation \minor{procedure} practical.}

  \item \rev{We introduce pseudo-off-policy RL and performance-based filtering, allowing the teacher to adapt to student behavior and provide high-quality demonstrations. This relaxes the reliance on oracle-level teachers assumed in prior KD methods.} 

  \item \minor{We validate HINT} consistently outperforms \rev{competitive} CTDE and \rev{KD} baselines with improvements of 60\% to 165\% in success rate, demonstrating enhanced coordination and robustness.
\end{itemize}

\section{Related Work}
\subsection{CTDE in Cooperative MARL}
CTDE has been widely adopted in cooperative MARL to reduce instability from non-stationarity by leveraging centralized value functions during training while enabling decentralized execution. These methods fall into off-policy~\cite{QMIX, qtran, maac, sqddpg, MADDPG} and on-policy categories~\cite{COMA, MAPPO, dae, HATRPO_HAPPO}, balancing sample efficiency and learning stability. For instance, QMIX~\cite{QMIX} and QTRAN~\cite{qtran} use experience replay buffers for scalability, while MAPPO~\cite{MAPPO} and DAE~\cite{dae} rely on trust-region updates to mitigate unstable gradients.

A persistent and fundamental challenge in CTDE is fairly assigning credit to agents according to their individual contributions to team rewards. Various strategies have been proposed to address credit assignment: value decomposition (QMIX~\cite{QMIX}, QTRAN~\cite{qtran}), attention-based critics (MAAC~\cite{maac}), and explicit fairness estimators like Shapley value (SQDDPG~\cite{sqddpg}) or counterfactual baselines (COMA~\cite{COMA}). More recent methods like HAPPO~\cite{HATRPO_HAPPO} and DAE~\cite{dae} further refine gradient flows by decomposing joint advantages or leveraging difference rewards.

To supplement centralized training and enable adaptive coordination during decentralized execution, some CTDE methods include communication mechanisms. These include continuous channels (CommNet~\cite{commnet}), conditional activation (IC3Net~\cite{IC3Net}), and attention-based messaging (TarMAC~\cite{tarmac}, HetNet~\cite{HetNet1, hetnet2}), which allow agents to share contextual cues and prioritize relevant interactions.

Despite these advancements, \rev{CTDE methods continue to struggle with unstable training due to fundamental architectural constraints in shared-value decomposition and inter-agent coordination.} To overcome these challenges, we adopt a centralized hierarchical RL framework, which decomposes complexity and allows for coarse-to-fine temporal abstraction (Sec.~\ref{sec:hitmac}) and enable decentralized execution through knowledge distillation (Secs.~\ref{sec:hetnet} and ~\ref{sec:knowledge_distillation}).

\subsection{Multi-Agent Learning with KD} Recently, KD has emerged in MARL as a promising method for transferring coordination strategies from centralized to decentralized agents. One of the main advantages of \rev{KD} is the provision of stable training signals by leveraging a teacher model with access to global states or joint observations, implicitly addressing challenges such as credit assignment and non-stationarity. This benefit, however, often assumes an oracle-like teacher that consistently offers optimal guidance.

Distillation strategies vary by (1) what is distilled (e.g., value~\cite{ctds, ddn} vs. policy~\cite{cesma}), (2) who teaches (e.g., planners~\cite{primal, graph_mapf}, humans~\cite{mixture, human_ma_navigation}), and (3) how interactivity is handled (e.g., one-shot~\cite{madt, off_makd} vs. iterative~\cite{cesma, igm-da}). However, these methods often rely on static, oracle-level supervision, which is brittle in open-ended environments. Our empirical findings suggest that this assumption of optimality breaks down in \rev{complex, dynamic} cooperative tasks, particularly when teachers are likely to encounter unfamiliar states (Sec.~\ref{sec:suboptimality}). To overcome this limitation, we enable adaptive supervision through a \rev{hierarchical} teacher that continuously refines its policy based on student behaviors (Sec.\ref{method:pseudoRL}) and applies a performance-based filtering to selectively distill high-quality demonstrations (Sec.\ref{method:metaheuristic}).

\section{Preliminaries}

\subsection{Hierarchical Target-oriented Multi-Agent 
Coordination} 
\label{sec:hitmac}
We employ hierarchical target-oriented multi-agent 
coordination (HiT-MAC)~\cite{HiT-MAC} as our centralized teacher because of its robust performance \rev{in dynamic, complex environments}. Specifically, HiT-MAC facilitates efficient multi-agent training through hierarchical agent-target structures. In Fig.~\ref{fig:overview}b, a high-level coordinator assigns subgoals based on joint observations, while each agent acts as a low-level executor, selecting primitive actions using its local observation and assigned subgoal.

Technically, HiT-MAC operates under a centralized training and centralized execution (CTCE) paradigm, leveraging a self-attention mechanism~\cite{attention} over joint observations to capture inter-agent dependencies and evaluate the relative importance of agent-target pairs. This design supports effective credit assignment via Shapley value approximation, quantifying each pair’s contribution to overall performance. \minor{Through this mechanism, HiT-MAC not only resolves challenging credit assignment problems in CTDE methods but also establishes itself as a reliable, high-performing teacher for decentralized students.} Empirically, HiT-MAC achieves over 80\% success rates in our challenging benchmark scenarios, outperforming tested \rev{CTDE and KD} baselines. To accommodate agent heterogeneity, we extend the original design with class-specific encoders that handle diverse observation modalities and information access. Further design details are provided in Appendix~\ref{Appendix:teacher and student models}.

\subsection{Heterogeneous Policy Network (HetNet)}
\label{sec:hetnet}
To support decentralized execution among heterogeneous agents, we adopt HetNet~\cite{hetnet2, HetNet1} as our student model. HetNet is specifically designed to support decentralized coordination by modeling agent-specific observation and action modalities.

Fig.~\ref{fig:overview}c shows that each agent processes its local observations through a preprocessing module, followed by an LSTM~\cite{lstm}, which helps mitigate partial observability by retaining relevant historical information. To enable structured communication, we employ heterogeneous graph attention (HetGAT) layers that assign attention weights conditioned on both agent types and their relational context. This mechanism enables agents to selectively attend to relevant teammates and share task-specific information. Finally, a decoder integrates the local features with aggregated graph messages to produce the action probability. \rev{Further architectural details are provided in Appendix~\ref{Appendix:teacher and student models}.}



\subsection{Notation for Teacher and Student Policies}
The teacher policy $\pi_T$ is defined over the joint observations 
$\bar{o} = (o^1, \ldots, o^n)$ and joint actions 
$\bar{a} = (a^1, \ldots, a^n)$ as
\begin{equation}
\pi_T(\bar{a}|\bar{o}) 
= \sum_{\bar{g}} \pi_T^H(\bar{g}|\bar{o}) 
  \prod_{i=1}^{n} \pi_T^{L_i}(a^i|o^i, g^i)
\end{equation}
where $\pi_T^H$ is a high-level policy that assigns subgoals 
$\bar{g} = (g^1, \ldots, g^n)$ based on the joint observations $\bar{o}$, 
and $\pi_T^{L_i}$ is a low-level policy for agent $i$, selecting 
action $a^i$ using its observation $o^i$ and the assigned subgoal $g^i$. 
The overall teacher policy $\pi_T$ and value function $V^{\pi_T}$ 
are parameterized by $\theta$ and $\psi$, respectively.

The joint student policy $\pi_S$ is defined over joint 
observations $\bar{o} = (o^1, \ldots, o^n)$ and joint actions 
$\bar{a} = (a^1, \ldots, a^n)$ as
\begin{equation}
\pi_S(\bar{a} \mid \bar{o}) 
= \prod_{i=1}^{n} \pi_{S_i}(a^i \mid o^i)
\end{equation}
where each $\pi_{S_i}$ is a decentralized policy that selects 
$a^i$ based on local observation $o^i$. 
The policy $\pi_S$ is parameterized by $\phi$.

\section{Student State Distribution When Scaling Environments}
\label{sec:suboptimality}

\begin{figure}[h]
\centering

  \includegraphics[width=0.46\textwidth]{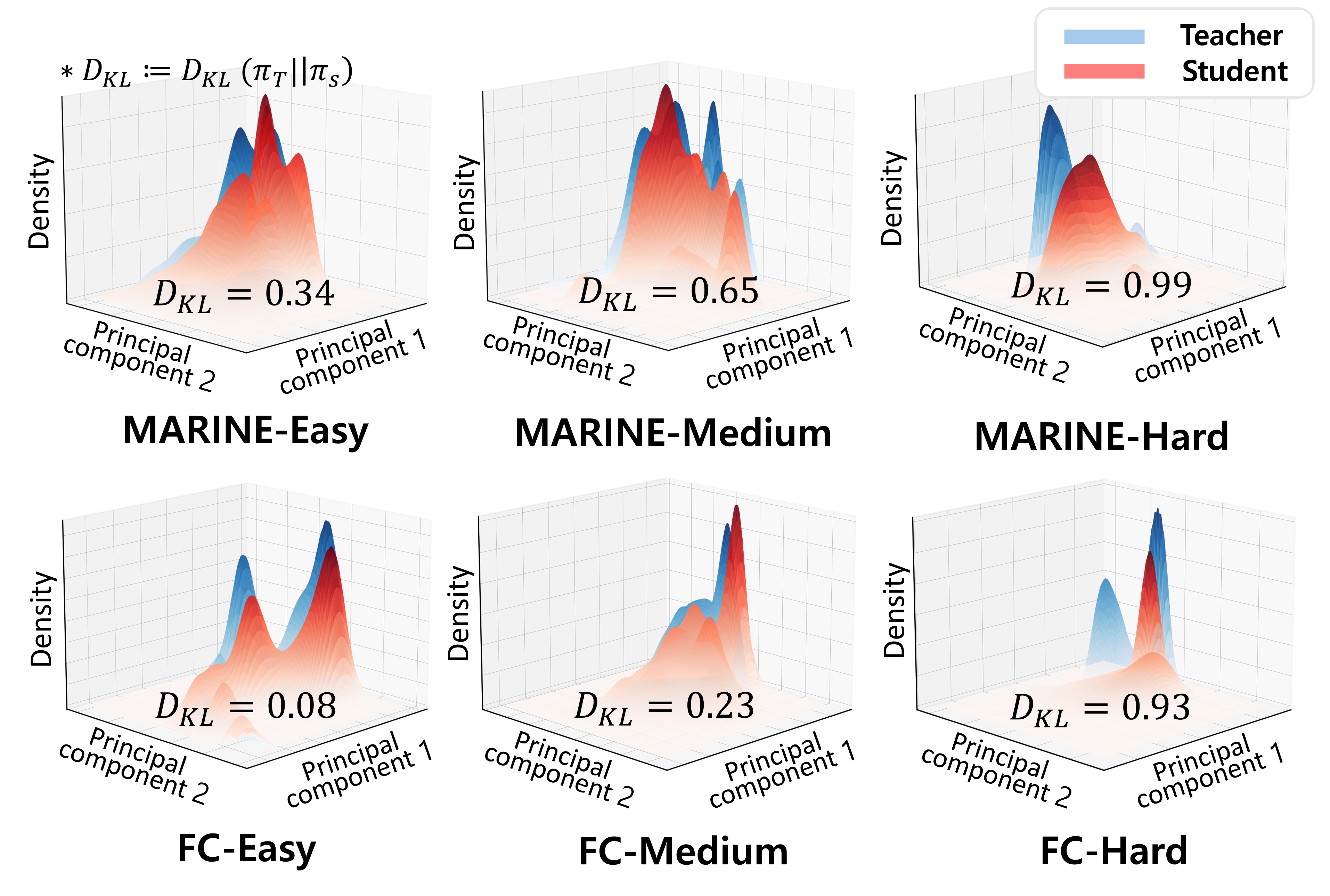}
  \vspace{-7pt}
  \caption{Comparison of teacher (blue) and student (red) state distributions projected into a shared latent space for MARINE and FC. Each setting corresponds to student rollouts with a 10–30\% success rate. As task complexity increases, the gap between student and teacher distributions widens (measured by KL-divergence), indicating that teachers are increasingly exposed to out-of-distribution (OOD) states.}
  \vspace{-7pt}
  \label{fig:suboptimality}
\end{figure}
This section investigates how student state distributions shift as the environment scales and team size grows. Focusing on student rollouts with a 10–30\% success rate, we compare their trajectories with teacher demonstrations by projecting both into a shared latent space (see Fig.~\ref{fig:suboptimality}). Here, PCA~\cite{abdi2010principal} is utilized to visualize differences in a common latent space. In smaller settings, student distributions generally align with those of the teacher, suggesting that the teacher may still provide high-quality demonstrations. However, in more complex environments with larger agent teams, students diverge (e.g., higher KL-divergence) and result in more OOD states, diminishing the teacher's ability to offer consistent and optimal guidance. To address this challenge, we propose two key components: pseudo off-policy RL and performance-based filtering (Secs.~\ref{method:pseudoRL} and~\ref{method:metaheuristic}). Their effectiveness is evaluated in Secs.~\ref{sec:benchmark} and~\ref{sec:ablation}.\footnote{\rev{Additonal examples of suboptimal demonstrations and the procedure for Fig.~\ref{fig:suboptimality} are given in Appendix~\ref{Appendix:suboptimal data}, with environment details in Sec.~\ref{environment} and Appendix~\ref{Appendix:env details}.}}

\section{Method}
In this section, we propose \rev{HINT}, a framework for interactive knowledge distillation designed to support robust multi-agent learning (see Fig.~\ref{fig:overview}a). HINT integrates three components: (1) knowledge distillation, (2) pseudo off-policy RL, and (3) performance-based filtering. At its core, the teacher refines its policy using student trajectories to provide more informative guidance. Meanwhile, the student runs its policies and selectively queries the teacher for improved actions, ensuring that only outcome-relevant feedback is retained. This cycle repeats until student policies converge. \rev{Please note that HINT adopts a multi-threaded structure throughout all submodules to enhance computational efficiency.}

\subsection{Knowledge Distillation}
\label{sec:knowledge_distillation}
Compared to prior \rev{KD} methods in \rev{MARL}, we provide more \rev{efficient and robust} demonstrations using a centralized, hierarchical teacher, because it enables \rev{systematic} decomposition of decision-making, which improves tractability in \rev{complex} multi-agent tasks. Since the teacher can access all agents’ observations and infer the global context, it can efficiently address credit assignment among agents. However, to enable decentralized execution of agents, transferring knowledge from the teacher to student is necessary. \rev{The overall procedure is summarized in Alg.~\ref{alg:knowledge_distillation}.} Note that while we follow the typical KD procedure, our approach is distinct in that the dataset $\mathcal{D}$ is continuously augmented with an adaptive teacher, thereby providing richer guidance to the students.

Our policy distillation objective minimizes the difference between the teacher policy $\pi_T$ and student policy $\pi_S$ by combining KL-divergence with an entropy regularization term to encourage both imitation and policy diversity. The loss is weighted by a tunable coefficient $\alpha$, as shown in Eq.~\ref{eq:student_loss}. 
\begin{align}
\label{eq:student_loss}
    \mathcal{L}_\phi = 
    \mathbb{E}_{(\bar{o}_t,\bar{a}_t) \sim \pi_{T}} \Bigl[
        & \bigl(\log \pi_{T}(\bar{a}_t \mid \bar{o}_t) 
        - \log \pi_{S}(\bar{a}_t \mid \bar{o}_t)\bigr) \nonumber \\
        &\quad+ \alpha \mathcal{H}(\pi_{S}(\cdot \mid \bar{o}_t))
    \Bigr]
\end{align}
To ensure stable training, samples are stored in a replay buffer and updated using batch-based gradient optimization. We empirically found this formulation yields stable gradients \rev{across tested benchmarks} (see Appendix~\ref{Appendix:ablation study}).


\subsection{Pseudo Off-Policy Reinforcement Learning}
\label{method:pseudoRL}
\begin{figure}[t]
\centering
\includegraphics*[width=0.95\linewidth]{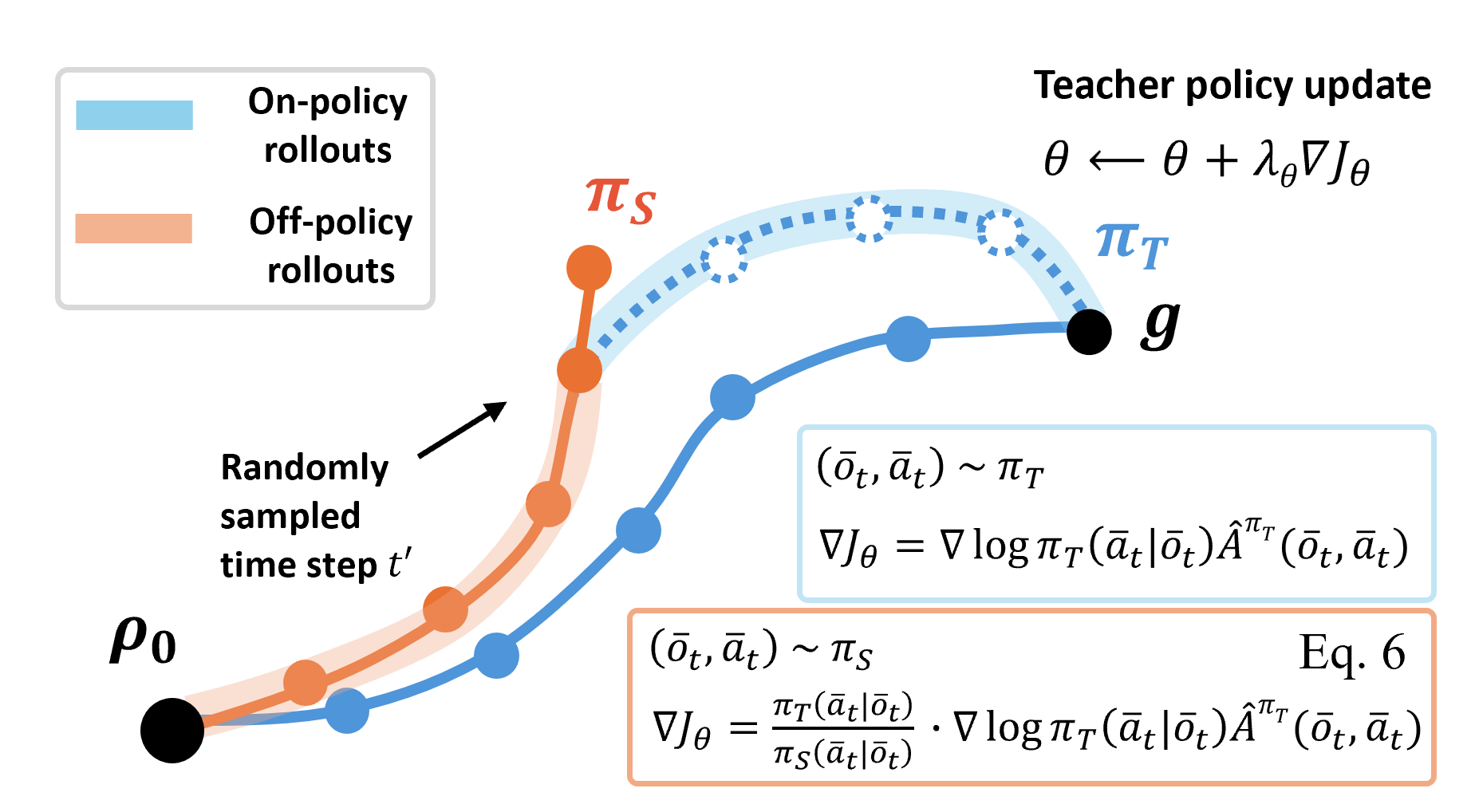}
\vspace{-10pt}
\caption{Teacher policy ($\pi_T$) is refined using its on-policy rollouts (shaded blue) and off-policy rollouts (shaded orange) from the student ($\pi_S$), with policy gradients corrected via importance sampling.}
\label{fig:pseudo_rl}
\end{figure}

Our pseudo off-policy RL presents a new perspective on how the teacher and student co-evolve in multi-agent learning. The overall procedure is shown in Alg.~\ref{alg:pseudo_off_policy}. Rather than treating student and teacher data separately, we combine them within a single trajectory: the student explores initially, then the teacher completes the path (lines 5-8). This joint rollout structure (Fig.\ref{fig:pseudo_rl}) bridges the gap between teacher and student, allowing the teacher to learn not only what to do, but how to recover from student mistakes (lines 11-13). This approach yields a more resilient teacher and facilitates student learning through richer, context-aware guidance—especially in \rev{complex} settings where students often deviate from training distributions and reactive corrections alone are insufficient.
\SetAlgoSkip{}
\begin{algorithm}[t]
\LinesNumbered
\footnotesize
\SetCommentSty{textnormal}
\setlength{\algomargin}{0.5em}
\SetInd{0.5em}{0.5em}
\renewcommand{\baselinestretch}{0.9}\selectfont
\caption{Knowledge Distillation}
\label{alg:knowledge_distillation}
\KwIn{Dataset $\mathcal{D} = \left\{ \tau = (\bar{o}_t, \bar{a}_t)_{t=1}^H \right\}$, Buffer $\mathcal{B}$
}
\Initialize{$\phi$ \textnormal{as the parameters of} $\pi_{S}$}

\For{$\tau$ \textnormal{in} $\mathcal{D}$}
{\For{$t = 1$ \KwTo $H$}
{
    Compute $\log \pi_S(\bar{a}_t|\bar{o}_t)$, and $\mathcal{H}(\pi_S(\cdot \mid \bar{o}_t))$ and store in $\mathcal{B}$\;
    \If{$\mathcal{B}$ \textnormal{is full}}{
    Compute loss $\mathcal{L}_{\phi}$ using Eq.~\ref{eq:student_loss}\tcp*{KL-div. with entropy regularization}
    $\phi \leftarrow \phi - \lambda_\phi \nabla_\phi \mathcal{L}_{\phi}$\;
    Clear buffer $\mathcal{B}$\;}
}
}
\end{algorithm}
\begin{algorithm}[t]
\LinesNumbered
\footnotesize
\SetCommentSty{textnormal}
\setlength{\algomargin}{0.5em}
\SetInd{0.5em}{0.5em}
\renewcommand{\baselinestretch}{0.92}\selectfont
\caption{Pseudo Off-Policy RL}
\label{alg:pseudo_off_policy}
\KwIn{Teacher $\pi_T$, Student $\pi_S$, Buffer $\mathcal{B}$, Episodes $N_{pseudo}$}
Freeze $\pi_S$\;
\For{$e = 1$ \KwTo $N_{pseudo}$}{
    Sample switch point $t' \sim \text{Uniform}(1, H)$\;
    \For{$t = 0$ \KwTo $H$}{
        \eIf{$t \leq t'$}{
                $\bar{a}_t \sim \pi_S(\cdot \mid \bar{o}_t)$\tcp*{Student explores}
            }{
                $\bar{a}_t \sim \pi_T(\cdot \mid \bar{o}_t)$\tcp*{Teacher resumes}
            }
        Store $(\bar{o}_t, \bar{a}_t)$ in $\mathcal{B}$\;
        \If{$\mathcal{B}$ \textnormal{contains} $n$ \textnormal{steps}}{
            Compute $v_t^{\pi_T}$ (Eq.~\ref{eq:v-trace_target})\tcp*{Correct off-policy}
            Compute losses $\mathcal{L}_{\psi}$, $J_{\theta}$ (Eqs.~\ref{eq:teacher_value_function},~\ref{eq:pseudo_policy_objective})\;
            $\psi \leftarrow \psi - \lambda_\psi \nabla_\psi \mathcal{L}_\psi$, 
            $\theta \leftarrow \theta + \lambda_\theta \nabla_\theta J_\theta$\tcp*{Update teacher}
            Clear $\mathcal{B}$\;
        }
    }
}
\end{algorithm}
\setlength{\textfloatsep}{5pt}

To effectively leverage these hybrid trajectories, we require a stable way to update the teacher using off-policy data. To this end, we employ the V-trace method from IMPALA \cite{impala}, \minor{which has been shown to provide stable convergence through bounded off-policy corrections. This allows us} to estimate corrected value targets $v_t^{\pi_T}$ from a mixture of student and teacher trajectories (line 11).
\begin{equation}
\begin{aligned}
v^{\pi_T}_t = V^{\pi_T}(\bar{o}_t) + \sum_{j=t}^{t+n-1} \gamma^{j-t} 
\bigl( \prod_{i=t}^{j-1} c_i \bigr) \delta_jV
\label{eq:v-trace_target}
\end{aligned}
\end{equation}
where $\delta_jV = \rho_j \left( r_j + \gamma V^{\pi_T}(\bar{o}_{j+1}) - V^{\pi_T}(\bar{o}_j) \right)$ is a temporal difference error scaled by truncated importance weights, $\rho_j = \min\bigl(1, \tfrac{\pi_T(\bar{a}_j|\bar{o}_j)}{\mu(\bar{a}_j|\bar{o}_j)}\bigr)$
and $c_i = \min \bigl( 1, \tfrac{\pi_{T}(\bar{a}_i | \bar{o}_i))}{\mu(\bar{a}_i | \bar{o}_i))} \bigr)$. Here, $\mu$ denotes the behavior policy, either $\pi_T$ or $\pi_S$. Note that while we utilize the V-trace without modification, our novelty lies in its first application to address mismatched teacher-student distributions in multi-agent distillation setups.

We train the teacher's value function $V^{\pi_T}(\bar{o}_t)$ by minimizing the squared error between its predictions and the corrected V-trace targets in Eq.~\ref{eq:teacher_value_function} (line 12). This corrected value target $v_t^{\pi_T}$ helps stabilize training even when student trajectories deviate significantly from the teacher’s policy, which could otherwise lead to instability.
\begin{equation}
\begin{aligned}
\mathcal{L}_\psi = \mathbb{E}_{(\bar{o}_t, \bar{a}_t) \sim \{ \pi_S, \pi_T \}} 
\left[ \left( V^{\pi_T}(\bar{o}_t) - v^{\pi_T}_t \right)^2 \right]
\label{eq:teacher_value_function}
\end{aligned}
\end{equation}

Similarly, the policy update integrates trajectories from both teacher and student. For student-generated samples, we apply off-policy corrections using importance weighting, while teacher-generated samples remain on-policy. Note that the advantage estimate $\hat{A}^{\pi_T}(\bar{o}_t, \bar{a}_t)$ is computed using the V-trace target as $r_t + \gamma v_{t+1}^{\pi_T} - V^{\pi_T}(\bar{o}_t)$. The resulting policy objective is given in Eq. \ref{eq:pseudo_policy_objective} (line 12), where off-policy corrections with augmented advantage estimation provide a more robust and stable training signal, without being misled by off-policy drift.
\begin{equation}
\label{eq:pseudo_policy_objective}
\begin{aligned}
J_\theta&= \mathbb{E}_{(\bar{o}_t,\bar{a}_t)  \sim \pi_{S}} \Bigl[\frac{\pi_{T}(\bar{a}_t|\bar{o}_t)}{\pi_{S}(\bar{a}_t|\bar{o}_t)} \log \pi_{T}(\bar{a}_t|\bar{o}_t) \hat{A}^{\pi_{T}}(\bar{o}_t, \bar{a}_t) \Bigr]
\\&\quad+\mathbb{E}_{(\bar{o}_t,\bar{a}_t)  \sim \pi_{T}} \bigl[ \log \mathrm{\pi}_{T}(\bar{a}_t|\bar{o}_t) \hat{A}^{\mathrm{\pi}_{T}}(\bar{o}_t,\bar{a}_t) \bigr]
\end{aligned}
\end{equation}

The objectives in Eq. \ref{eq:teacher_value_function} and Eq. \ref{eq:pseudo_policy_objective} allow adaptive refinement of the teacher using both on-policy and off-policy data (line 10). This hybrid learning scheme enhances generalization while preserving \rev{policy consistency}. During this adaptive phase, the teacher’s low-level policy remains fixed, as it was pretrained on a sufficiently diverse distribution of states. This design choice simplifies the optimization process and focuses adaptation on strategic goal assignment.
\setlength{\textfloatsep}{7pt}
\begin{figure}[t]
\centering
\includegraphics*[width=0.82\linewidth]{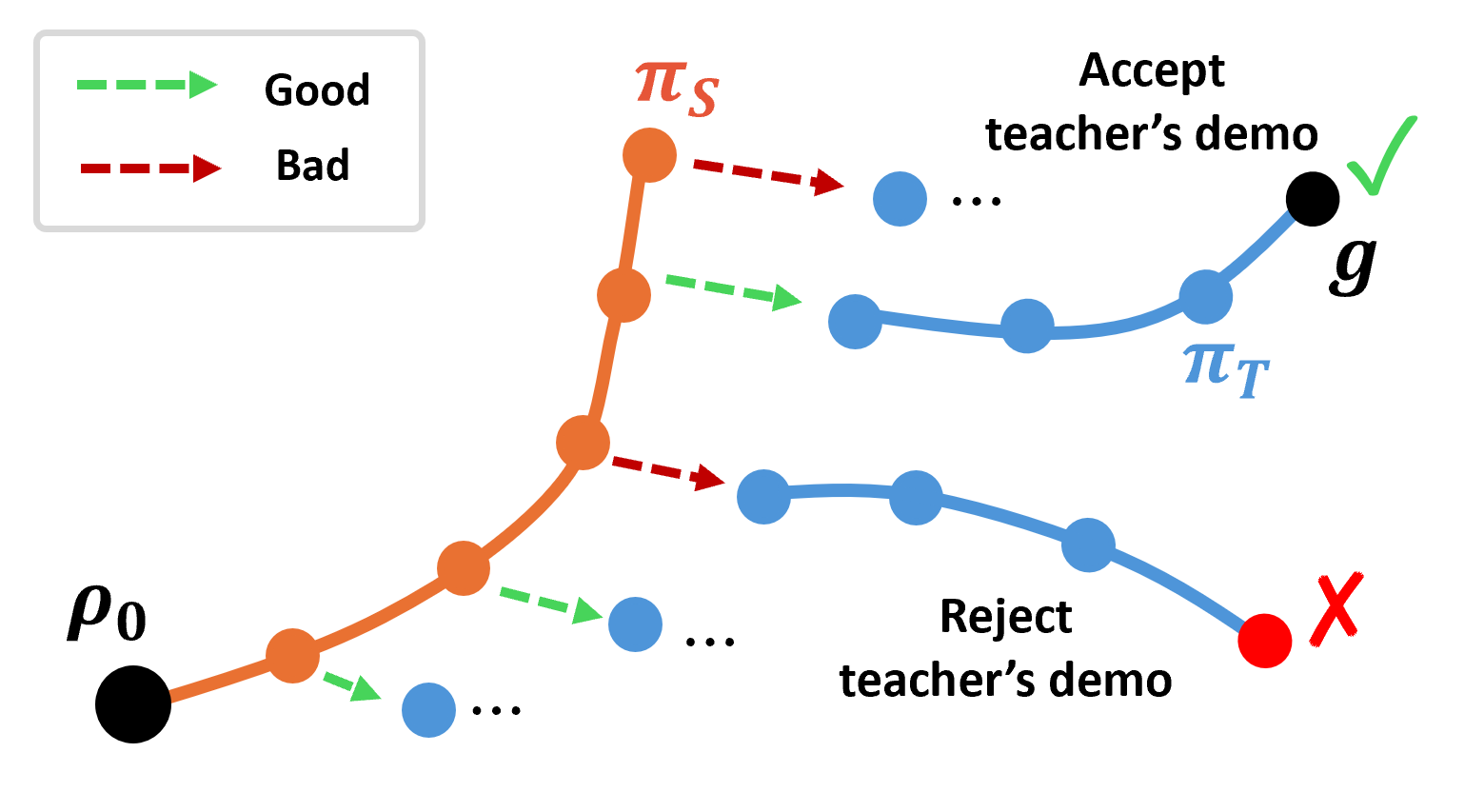}
\vspace{-10pt}
\caption{\minor{Performance-based filter is applied during dataset aggregation. High-quality teacher demonstrations are accepted (green, \ding{51}), while suboptimal ones are rejected (red, \ding{55}).}}
\label{fig:meta_filter}
\end{figure}

\subsection{DAgger with Performance-Based Filtering}
\label{method:metaheuristic}
Our approach builds on the standard DAgger (Dataset Aggregation)~\cite{dagger}, which addresses distributional shift by iteratively training student policies on an expanding dataset of observation-action pairs. We collect these pairs by running the student policies in the environment and querying the expert policy at the states they visit. Training on this aggregated dataset allows the student to learn from a data distribution similar to what they will see at test time. However, DAgger usually assumes access to a reliable teacher. In practice, especially in \rev{open-ended} environments (Fig.~\ref{fig:suboptimality}), our hierarchical teacher policy may face unfamiliar or unseen states \new{during interactive queries}, leading to inconsistent demonstrations and \rev{compounding observation mismatches between the centralized teacher and decentralized students.}

\begin{algorithm}[t]
\LinesNumbered
\footnotesize
\SetCommentSty{textnormal}
\setlength{\algomargin}{0.5em}
\SetInd{0.5em}{0.5em}
\renewcommand{\baselinestretch}{1.05}\selectfont
\caption{DAgger with Performance-Based Filter}
\label{meta}
\KwIn{Teacher $\pi_T$, Student $\pi_S$, Dataset $\mathcal{D}$, Total Episodes $N_{\text{query}}$}
\BlankLine
\For{$e = 1$ \KwTo $N_{\text{query}}$}{
    Initialize environment, set trajectory $\tau \leftarrow [\,]$\tcp*{Student explores}
    \For{$t = 1$ \KwTo $H$}{
        Execute student action $\bar{a}_t \sim \pi_S(\cdot | \bar{o}_t)$\; 
        Query teacher action $\bar{a}^*_t \sim \pi_T(\cdot | \bar{o}_t)$\tcp*{Teacher guides}
        Simulate teacher policy $\pi_T$ from $(\bar{o}_t, \bar{a}^*_t)$ to terminal state $\bar{o}_H$\;
        \If{\textnormal{terminal state} $\bar{o}_H$ \textnormal{satisfies success condition}}{
            Append $(\bar{o}_t, \bar{a}^*_t)$ to trajectory $\tau$ \tcp*{performance-based filter}
        }
    }
    Add $\tau$ to dataset $\mathcal{D}$\tcp*{Aggregate dataset}
}
\end{algorithm}

To address this challenge, we propose a performance-based filtering mechanism that operates during data aggregation to enhance the quality of expert actions. The overall procedure is described in Alg.~\ref{meta}. As in Fig.~\ref{fig:meta_filter}, when the student queries the teacher at each timestep \rev{(line 5)}, we simulate the teacher’s trajectory from that point to the end of the episode \rev{(line 6)}. Suppose the resulting terminal state meets the task-specific success criterion \rev{(line 7)}. In that case, we consider the initially queried action as a proxy for a good action (green) and include it in the dataset \rev{(lines 8 and 9)}. Otherwise, the expert action is discarded. While the teacher provides a heuristic proxy for optimal behavior, our mechanism adds a second layer of evaluation to assess the trustworthiness of this proxy by simulating its long-term consequences. 

This domain-agnostic filtering approach makes the data aggregation more robust to noisy or suboptimal demonstrations. By selectively incorporating only expert actions empirically linked to successful outcomes, the student policy is trained on higher-quality data. Also, to stabilize training and support the evolving nature of the teacher policy, we retain a fixed number of initial demonstrations and periodically sample recent interactions. While this strategy is similar to experience replay with curriculum bias, our method uniquely blends static expert supervision with adaptive feedback, providing both reliable foundations and fresh guidance that aligns with the student’s current capabilities.

\section{Results and Discussion}
\label{results and discussion]}

\begin{table*}[t]
\centering
\caption{Quantitative results for MARINE and FireCommander (FC) in the online CTDE benchmark, reported as the mean ($\pm$ standard deviation) across three random seeds. Here, \textit{WS} denotes warm starting RL with our teacher policy.}
\resizebox{0.96\linewidth}{!}{
\begin{tabular}{lcccccccc}
\toprule
\multirow{2}{*}{\textbf{Method}} & 
\multicolumn{2}{c}{\textbf{MARINE-Medium (10x10, 5 agents)}} & 
\multicolumn{2}{c}{\textbf{MARINE-Hard (20x20, 10 agents)}} & 
\multicolumn{2}{c}{\textbf{FC-Medium (10x10, 5 agents)}} & 
\multicolumn{2}{c}{\textbf{FC-Hard (21x21, 10 agents)}} \\
\cmidrule(lr){2-3}\cmidrule(lr){4-5}\cmidrule(lr){6-7}\cmidrule(lr){8-9}
 & Success Rate (\%)~$\uparrow$ & Steps Taken~$\downarrow$ 
 & Success Rate (\%)~$\uparrow$ & Steps Taken~$\downarrow$ 
 & Success Rate (\%)~$\uparrow$ & Steps Taken~$\downarrow$ 
 & Success Rate (\%)~$\uparrow$ & Steps Taken~$\downarrow$ \\
\midrule
MAPPO   & 40.67 $\pm$ 51.39 & 69.23 $\pm$ 37.31 & 0.67 $\pm$ 1.15 & 198.88 $\pm$ 1.94 & 6.00 $\pm$ 3.46 & 96.33 $\pm$ 1.19 & 4.00 $\pm$ 0.00 & 202.89 $\pm$ 0.86 \\
TarMAC   & 94.00 $\pm$ 10.39 & \textbf{25.84 $\pm$ 14.77} & 0.00 $\pm$ 0.00 & 200.0 $\pm$ 0.00 & 5.33 $\pm$ 5.03 & 96.89 $\pm$ 3.97 & 2.67 $\pm$ 3.06 & 207.51 $\pm$ 2.48 \\
IC3Net   & 88.00 $\pm$ 10.58 & 34.97 $\pm$ 17.99 & 0.00 $\pm$ 0.00 & 200.0 $\pm$ 0.00 & 6.67 $\pm$ 6.11 & 97.07 $\pm$ 2.79 & 2.67 $\pm$ 3.06 & 207.08 $\pm$ 2.68 \\
CommNet & 0.00 $\pm$ 0.00 & 100.00 $\pm$ 0.00 & 0.00 $\pm$ 0.00 & 200.0 $\pm$ 0.00 & 3.33 $\pm$ 3.06 & 97.81 $\pm$ 1.92 & 1.33 $\pm$ 2.31 & 207.96 $\pm$  1.77 \\
HAPPO     & 98.67 $\pm$ 2.31 & 27.81 $\pm$ 0.46 & 0.00 $\pm$ 0.00 & 200.0 $\pm$ 0.00 & 0.67 $\pm$ 1.15 & 99.35 $\pm$ 1.13 & 0.67 $\pm$ 1.15 & 208.61 $\pm$ 2.40 \\
\hspace{0.1em}$\rightarrow$ 25\% \textit{WS}     & - & - & 24.00 $\pm$ 3.46 & 169.94 $\pm$ 9.25 & - & - & - & - \\
\hspace{0.1em}$\rightarrow$ 50\% \textit{WS}     & - & - & 38.00 $\pm$ 6.93 & 154.73 $\pm$ 8.30 & - & - & - & - \\
HetNet   & 61.33 $\pm$ 53.27  & 60.33 $\pm$ 34.64 & 0.00 $\pm$ 0.00 & 200.0 $\pm$ 0.00 & 82.00 $\pm$ 14.00 & 49.43 $\pm$ 16.00 & 19.33 $\pm$ 7.02 & 177.93 $\pm$ 11.91 \\
\hspace{0.1em}$\rightarrow$ 25\% \textit{WS}     & - & - & - & - & - & - & 35.33 $\pm$ 15.28 & 152.35 $\pm$ 18.78 \\
\hspace{0.1em}$\rightarrow$ 50\% \textit{WS}     & - & - & - & - & - & - & 42.67 $\pm$ 38.28 & 139.01 $\pm$ 62.69 \\
\textbf{HINT (Ours)}          & \textbf{98.67 $\pm$ 1.15} & 26.49 $\pm$ 4.92 & \textbf{53.33 $\pm$ 9.02} & \textbf{143.28 $\pm$ 4.47} & \textbf{84.00 $\pm$ 2.00} & \textbf{47.95 $\pm$ 3.73} & \textbf{51.33 $\pm$ 5.03} & \textbf{136.40 $\pm$ 3.86} \\
\bottomrule
\end{tabular}
}
\label{table1}
\vspace{-8pt}
\end{table*}

\begin{figure*}[t]
  \centering
  \includegraphics[width=0.96\textwidth]{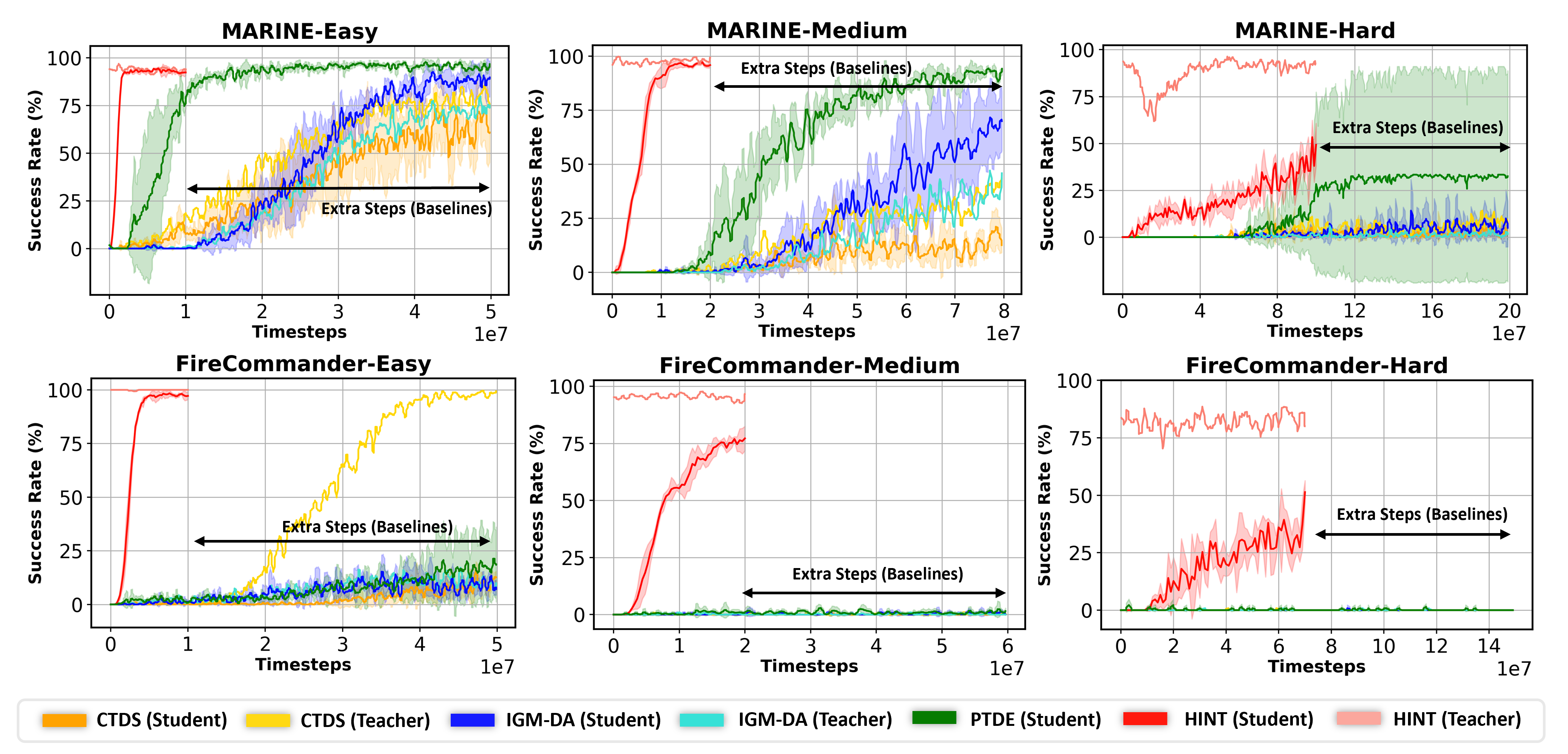}
  \vspace{-10pt}

  \caption{Learning curves for MARINE and FireCommander in the online KD benchmark, reported as the mean ($\pm$ standard deviation) across three random seeds and three difficulty settings. HINT consistently outperforms all baselines in both domains.}
  \label{fig:online_KD}
  \vspace{-10pt}
\end{figure*}

In this section, we evaluate the performance of \rev{HINT} by benchmarking against baselines (Sec.~\ref{sec:benchmark}) and validating the contribution of each submodule via ablation studies (Sec.~\ref{sec:ablation}). Detailed training setups, \rev{along with the complete benchmark and ablation results}, are provided in  Appendices~\ref{Appendix:training steup}, \ref{app:full_benchmark}, and~\ref{Appendix:ablation study}, respectively.

\subsection{Environments}
\label{environment}
We selected two multi-agent domains that are highly dynamic, partially observable, and heterogeneous. These domains represent real-world challenges requiring \rev{timely and effective} coordination, making them ideal for testing \rev{HINT}. Performance is measured using two key metrics: success rate, defined as the proportion of episodes completed successfully, and average steps taken, indicating the number of steps required to complete each task. For further details about the environments, refer to Appendix~\ref{Appendix:env details}

\textbf{\rev{MARINE~\cite{ras}}:} A maritime logistics environment where routing agents navigate dynamic ocean conditions, while logistic agents provide mid-sea refueling. The environment uses WaveWatch III forecast data~\cite{WW3}, introducing weather-induced uncertainty into planning and coordination. Routing agents must reach destinations before fuel depletion, requiring timely rendezvous with logistic agents. Three difficulty levels are used: easy (5×5, 2\textit{R}/1\textit{L}), medium (10×10, 3\textit{R}/2\textit{L}), and hard (20×20, 6\textit{R}/4\textit{L}), where \textit{R} and \textit{L} denote routing and logistic agents, respectively.

\textbf{FireCommander (FC)~\cite{HetNet1}:}  A grid-based wildfire environment inspired by FARSITE~\cite{FARSITE}, \rev{a widely used simulator that models spatial and temporal behavior of fires through differential equations. In FC, action agents are responsible for extinguishing fires but do not have perception capabilities, while perception agents monitor fire spread but cannot put out fires. This division of roles encourages collaboration in a stochastic environment shaped by dynamic fire propagation.} The environment features three difficulty levels: easy (5×5, 2\textit{A}/1\textit{P}), medium (10×10, 3\textit{A}/2\textit{P}), and hard (21×21, 6\textit{A}/4\textit{P}), where \textit{A} and \textit{P} denote action and perception agents, respectively.


\subsection{Benchmark Test}
\label{sec:benchmark}

\subsubsection{Online CTDE Baselines}
\label{sec:CTDE_test}
We compare our proposed method against competitive \rev{online CTDE methods, including two without communication (MAPPO~\cite{MAPPO} and HAPPO~\cite{HATRPO_HAPPO}), four with communication (TarMAC~\cite{tarmac}, IC3Net~\cite{IC3Net}, CommNet~\cite{commnet}, and HetNet~\cite{hetnet2}). For fairness, all baselines were trained with additional timesteps to match those used for training our teacher policy. To further validate the effectiveness of interactive distillation, we compare HINT with RL methods that leverage warm-starting on hard tasks. Here, warm-starting refers to initializing the RL baselines from our teacher policy via behavior cloning (BC), rather than training from scratch. Based on Table~\ref{table1}, HAPPO and HetNet serve as strong baselines in MARINE and FireCommander, respectively. Thus, we initialize these methods with 25\% and 50\% warm-starting using our teacher policy.}

Table~\ref{table1} shows that our proposed method consistently outperforms other CTDE baselines in \rev{both MARINE and FC across medium and hard tasks.} Specifically, a high success rate coupled with fewer steps taken indicates that our approach yields agents that are both reliable and efficient, even in challenging scenarios. \rev{Moreover, HINT achieves notable improvements over warm-started HAPPO in the MARINE-Hard task. We suspect that MARINE’s terminal condition—fuel depletion of routing agents—hinders exploration and leads agents to become easily trapped in local minima, even with warm-starting. In the FC-Hard task, HINT outperforms warm-started HetNet by 20-45\%, while also exhibiting a much smaller standard deviation, reflecting a more stable training process.}

\rev{These improvements can be attributed to the structured communication in our framework. However, compared to HetNet and other warm-started variants, which employ the same student policy as HINT or are initialized with the teacher's demonstrations, we can observe how stable training signals and adaptive teacher's guidance contribute to final performance. This informative guidance comes from our centralized hierarchical teacher and an adaptive mechanism guided by pseudo off-policy RL and performance-filtered data. Together, these components enable robust policy learning under partially observable environments and OOD states, while also reducing observation mismatches between teacher and students. \minor{Please note that all methods were trained with an equal number of samples to ensure fairness in comparison.} For the full comparisons and training curves, find Appendix~\ref{app:full_benchmark}.}

\subsubsection{Online KD Baselines}
\label{sec:benchmark2}
\rev{Here, we evaluate HINT against three online knowledge distillation (KD) methods: 
\begin{itemize}
    \item[1.]  CTDS~\cite{ctds} - It employs a centralized teacher that has access to global information and trains decentralized students through imitation learning.
    \item[2.] IGM-DA~\cite{igm-da} - Similar to CTDS, but the centralized teacher is trained on data collected by students, augmented with global information, and distills knowledge back to students via DAgger.
    \item[3.] PTDE~\cite{ptde} - Global Information Personalization (GIP) module is introduced as a teacher network to distill agent-personalized global knowledge into the agent’s local information.
\end{itemize}}
All KD baselines are designed for value-based MARL. Hence, we adopt QMiX~\cite{QMIX} as the base policy for both teacher and student, as it achieved the best performance in the original KD studies. To mitigate convergence issues arising from the co-evolution of teacher and student networks, we train the KD baselines for more timesteps than the CTDE baselines.

\rev{Fig.~\ref{fig:online_KD} shows that HINT outperforms other KD baselines in both MARINE and FC environments. In FC, since QMiX lacks a communication module, KD baselines struggle to achieve good performance. Nonetheless, the poor performance of CTDS and IGM-DA teachers, even with access to global information, underscores the robustness of our hierarchical teacher in more challenging settings. (Since the teacher in PTDE functions as an auxiliary module rather than a policy, its performance is not shown in Fig.~\ref{fig:online_KD}.)} \rev{In MARINE, only PTDE achieves performance comparable to HINT in easy and medium tasks, but HINT surpasses PTDE by 60\% on hard tasks. Similar to Sec.~\ref{sec:CTDE_test}, HINT's success is attributed to its structured communication, but the results support the importance of an adaptive hierarchical teacher in our entire pipeline. For complete quantitative results, find Appendix~\ref{app:full_benchmark}.}


\subsection{Ablation Study}
\label{sec:ablation}

\subsubsection{Effect of Key Modules}
\label{sec:ablation1}
To evaluate the impact of each key component in \rev{HINT}, we conduct an ablation study comparing the full model against three variants: one without pseudo off-policy RL, one without \rev{performance-based} filtering, and one with both components removed, effectively reducing the method to standard DAgger. Throughout training, both the teacher and student policies are evaluated based on success rate and average steps taken. For the teacher, we also report the suboptimal demonstration rate, which serves as a proxy for \rev{guidance} quality by measuring whether the trajectory generated by the teacher, after responding to a student query, leads to a successful outcome from the query point onward.

\begin{figure}[t]
  \centering
  \includegraphics[width=0.485\textwidth]{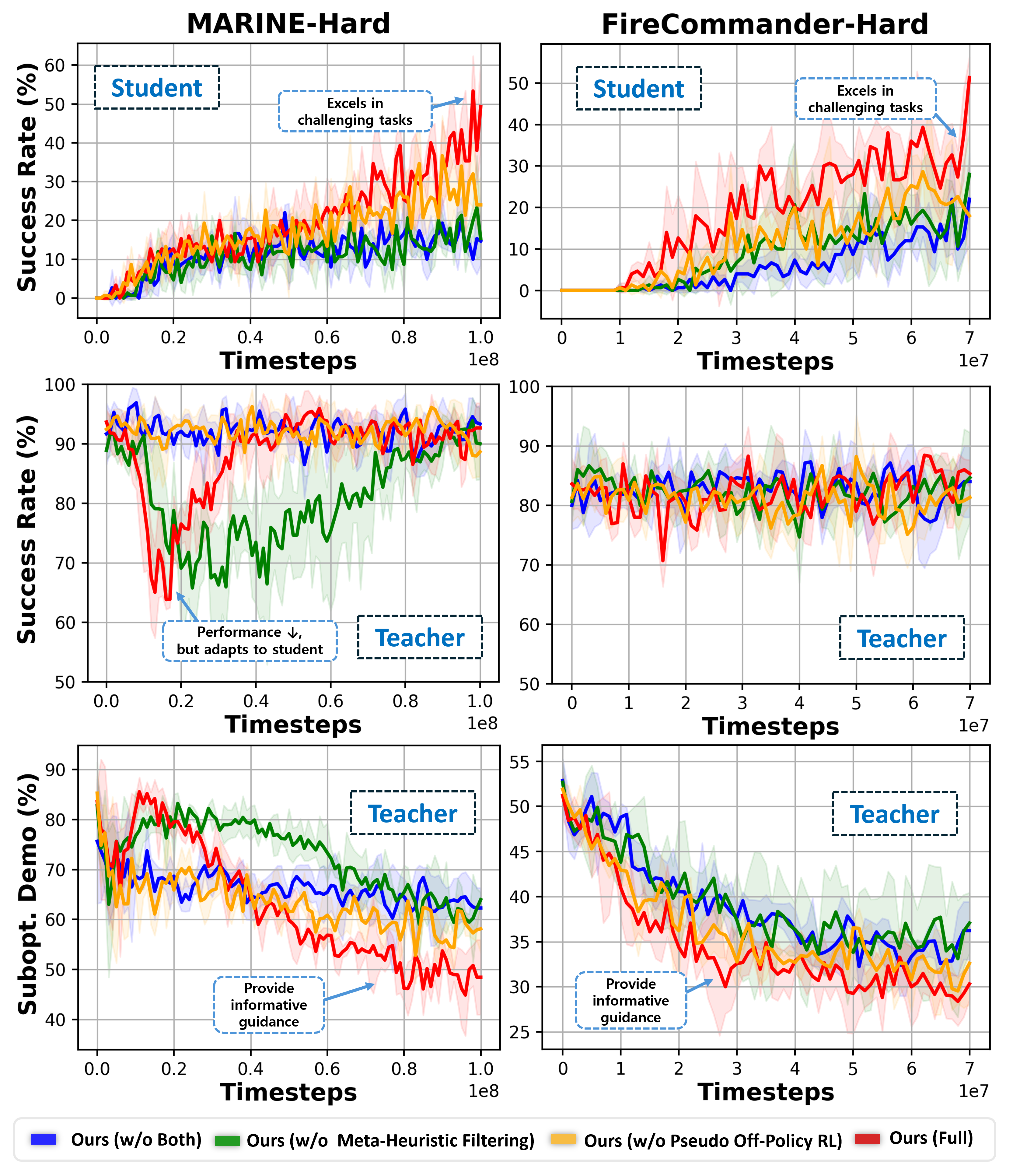}
  \caption{Ablation results of key modules on MARINE-Hard and FC-Hard. Each row corresponds to the student's success rate, teacher's success rate, and teacher's suboptimality demo rate, reported as the mean (± standard deviation) over three random seeds.}
  \label{fig:ablation_key}

\end{figure}

Fig.~\ref{fig:ablation_key} shows that the full model consistently outperforms all ablated variants \rev{with notable gains in both MARINE-Hard and FC-Hard tasks. Especially, we can observe that our full model shows a lower suboptimal demo rate compared to other variants, leading to higher student performance. Note that in MARINE-Hard condition,} the teacher's initial performance may decline as it adjusts to a variety of new and unfamiliar student queries. However, this leads to fewer suboptimal demonstrations, which in turn produces more informative guidance. Together, these findings confirm that HINT’s adaptability and filtering mechanisms are essential to enabling robust, high-quality policy learning. For complete ablation results, please find Appendix~\ref{Appendix:ablation study}.






\subsubsection{Student Structure}
\label{sec:ablation3}

To better understand the contributions of different architectural components in our student model, we conducted ablation studies on MARINE-Medium and FC-Medium tasks. Our proposed student model, HetNet, integrates two key modules: LSTM to address partial observability, and HetGAT mechanism for structured inter-agent communication. We compared HetNet against two ablated variants:
\begin{itemize}
  \setlength{\topsep}{3pt}
  \setlength{\itemsep}{0.5pt}
  \setlength{\parsep}{0.1pt}
    \item LSTM-Only: student model with LSTM (i.e., no communication)
    \item LSTM+GNN: student model with LSTM and GNN~\cite{graph_mapf}
\end{itemize}
As shown in Table~\ref{tab:student_ablation}, HetNet consistently outperforms both ablated baselines, with the largest gains in FC-Medium—improving the success rate by 84 points compared to LSTM-only, and by 18 points compared to LSTM+GNN. In this environment, structured communication is essential due to the limited sensing capabilities of action agents. The performance gap between the LSTM+GNN and HetNet highlights the value of heterogeneous attention in our architecture.

\subsubsection{Sensitivity Analysis}
\rev{We analyzed four key hyperparameters in FC-Medium task: $\alpha$ (student's entropy coefficient), $lr$ (learning rate), $N_{\text{pseudo}}$ (episodes for teacher training per epoch), and $N_{\text{query}}$ (episodes for expert queries per epoch). As shown in Table~\ref{tab:hyperparams}, HINT remains stable across a wide range of settings \minor{(-2.3\% to +6.3\% in success rate)}, with higher values generally yielding modest gains from additional training data and regularization. Overall, these results highlight HINT’s robustness and clarify how individual hyperparameters influence performance.}

\subsection{Limitations and Future Works}
\label{sec:limit, broader}
Since our teacher policy follows the CTCE paradigm, its scalability is fundamentally limited. Nevertheless, we choose a fully centralized teacher due to its high performance and try to improve scalability via hierarchical RL. \minor{In practice, HINT scales to 20×20 grid environments with up to 10 agents, which may appear modest in scale but remains challenging when jointly learning a communication protocol.} Furthermore, our framework includes more diverse modules than other CTDE and KD baselines, which may result in higher computational costs. However, we mitigate this by employing a multi-threaded structure across the entire pipeline, and we found that the overall computation time remains comparable to competitive baselines \minor{(e.g., HINT: 146 hours vs. PTDE: 161 hours on MARINE-Hard; HINT: 90 hours vs. HetNet: 179 hours on FC-Hard). Further details are provided in Appendix~\ref{app:time_analysis}.} Finally, HINT relies on human expertise to define appropriate hierarchical structures. While this design provides practical flexibility and is easy to implement, fully end-to-end training could yield more adaptive behaviors and reveal emergent capabilities. Future work could explore automating hierarchy construction through macro-action-based skill discovery and developing credit assignment mechanisms that operate asynchronously across agents based on the extracted skills.

\begin{table}[t]
\centering
\caption{\rev{Ablation results of student structure on MARINE-Medium and FC-Medium, reported as the mean ($\pm$ standard deviation) across three random seeds.}}
\label{tab:student_ablation}
\renewcommand{\arraystretch}{1.1}
\resizebox{\columnwidth}{!}{%
\begin{tabular}{lcccc}
\toprule
\multirow{2}{*}{\textbf{Method}} & \multicolumn{2}{c}{\textbf{MARINE-Medium}} & \multicolumn{2}{c}{\textbf{FC-Medium}} \\
\cmidrule(lr){2-3} \cmidrule(lr){4-5}
 & \shortstack{Success\\Rate (\%) $\uparrow$} & \shortstack{Steps\\Taken $\downarrow$} 
 & \shortstack{Success\\Rate (\%) $\uparrow$} & \shortstack{Steps\\Taken $\downarrow$} \\
\midrule
Ours (LSTM) & 92.67 $\pm$ 3.06 & 36.28 $\pm$ 6.31 & 0.00 $\pm$ 0.00 & 100.00 $\pm$ 0.00 \\
Ours (LSTM + GNN) & 98.00 $\pm$ 2.00 & 27.80 $\pm$ 0.97 & 66.00 $\pm$ 8.72 & 56.91 $\pm$ 6.17 \\
Ours (HetNet) & \textbf{98.67 $\pm$ 1.15} & \textbf{26.49 $\pm$ 4.92} & \textbf{84.00 $\pm$ 2.00} & \textbf{47.95 $\pm$ 3.73} \\
\bottomrule
\end{tabular}%
}
\end{table}

\begin{table}[t]
\centering
\caption{\rev{Performance comparison across different hyperparameters in FC-Medium, reported as the mean ($\pm$ standard deviation) across three random seeds.}}
\label{tab:hyperparams}
\renewcommand{\arraystretch}{1.0}
\resizebox{\columnwidth}{!}{
\begin{tabular}{ccccccc}
\toprule
\textbf{Hyper} & \multicolumn{3}{c}{$\alpha$} & \multicolumn{3}{c}{$\textit{lr}$} \\
\cmidrule(lr){2-4} \cmidrule(lr){5-7}
 \textbf{Parameter}& 0.01 (ours) & 0.02 & 0.05 & 1e-4 (ours) & 2e-4 & 5e-4 \\
\midrule
\makecell{Success\\Rate (\%) $\uparrow$}   & 84.00 $\pm$ 2.00 & 82.00 $\pm$ 8.00 & \textbf{89.33 $\pm$ 4.62} & 84.00 $\pm$ 2.00 & 86.00 $\pm$ 2.00 & \textbf{89.33 $\pm$ 4.62} \\
\makecell{Steps\\Taken $\downarrow$}   & 47.95 $\pm$ 3.73 & 47.27 $\pm$ 3.95 & \textbf{44.05 $\pm$ 5.39} & 47.95 $\pm$ 3.73 & 44.62 $\pm$ 2.24 & \textbf{44.05 $\pm$ 5.39} \\
\bottomrule
\end{tabular}
}

\vspace{0.5em}

\resizebox{\columnwidth}{!}{
\begin{tabular}{ccccccc}
\toprule
\textbf{Hyper} & \multicolumn{3}{c}{$N_{\text{pseudo}}$} & \multicolumn{3}{c}{$N_{\text{query}}$} \\
\cmidrule(lr){2-4} \cmidrule(lr){5-7}
  \textbf{Parameter}& 10 & 20 (ours) & 40 & 50 (ours) & 75 & 100 \\
\midrule
\makecell{Success\\Rate (\%) $\uparrow$}  & 84.67 $\pm$ 2.31 & 84.00 $\pm$ 2.00 & \textbf{85.33 $\pm$ 7.57} & 84.00 $\pm$ 2.00 & 85.33 $\pm$ 7.02 & \textbf{86.67 $\pm$ 8.08} \\
\makecell{Steps\\Taken $\downarrow$}   & 48.35 $\pm$ 5.98 & 47.95 $\pm$ 3.73 & \textbf{46.88 $\pm$ 6.41} & 47.95 $\pm$ 3.73 & 47.51 $\pm$ 7.36 & \textbf{44.66 $\pm$ 7.91} \\
\bottomrule
\end{tabular}
}
\end{table}
\section{Conclusion}
We introduced \rev{HINT}, a robust multi-agent learning framework leveraging \rev{hierarchical teacher-based adaptive} knowledge distillation. By integrating pseudo off-policy RL and performance-based filtering, our approach enables adaptive and robust knowledge transfer in \rev{dynamic, complex} cooperative tasks. Empirical evaluations demonstrate that \rev{HINT} consistently achieves robust performance even as environment complexity and team size increase. Our findings highlight the critical importance of addressing teacher policy suboptimality, thereby paving the way for future research on \rev{adaptive and robust} multi-agent systems.


\section*{Acknowledgments}
This work was supported by the Naval Research Laboratory under Grants N00173-21-1-G009 and N00173-25-1-0050 as well as MIT Lincoln Laboratory.

\bibliographystyle{plainnat}
\bibliography{bibliography}


\newpage
\appendices
\section{Environment Details}
\label{Appendix:env details}
{
\renewcommand{\arraystretch}{0.9}
\begin{table*}[t]
\caption{Environment configuration details for MARINE and FireCommander (FC).}
\vspace{-10pt}
\centering
\footnotesize
\resizebox{0.97\linewidth}{!}{
\begin{tabular}{c c c c c c c c c c}
\toprule
\textbf{Domain} & \textbf{Scenarios}& \textbf{Dimension} & \textbf{FOV} & \textbf{\makecell{\# Routing\\/Perception}} & \textbf{\makecell{\# Logistics\\/Action}} & \textbf{\makecell{\# Destination\\/Initial fire}} & \textbf{Max steps} & \textbf{\makecell{Initial Fuel\\(for routing)}} & \textbf{\makecell{Subarea Size\\(for teacher)}}\\

\midrule
MARINE & Easy   & 5x5 & 3x3  & 2 & 1 & 1 & 50 &5 &- \\
~   & Medium & 10x10 & 3x3  & 3 & 2 & 1 & 100&10 & - \\
~   & Hard   & 20x20 & 5x5  & 6 & 4 & 1 & 200&20 & - \\
\midrule
FC  & Easy   & 5x5 & 3x3  & 2 & 1 & 1 & 50 & - & 5x5 \\
~   & Medium & 10x10 & 3x3  & 3 & 2 & 1 & 100 & - & 5x5 \\
~   & Hard   & 21x21 & 5x5  & 6 & 4 & 1 & 210 & - & 7x7 \\
\bottomrule
\end{tabular}

}
\label{table:env_details}
\end{table*}
}

\begin{table*}[t]
\caption{Pre-training setup for teacher in MARINE and FireCommander (FC). \textit{H} and \textit{L} stand for high-level and low-level.}
\vspace{-10pt}
\centering
\footnotesize
\resizebox{0.97\linewidth}{!}{
\begin{tabular}{c c c c c c c c c c}
\toprule
\textbf{Domain} & \textbf{Scenarios}& \textbf{\makecell{Timesteps\\ (\textit{H}/\textit{L})}} & \textbf{\makecell{Learning\\ Rate}} & \textbf{\makecell{Entropy\\ Coefficient}} & \textbf{\makecell{Discount\\Factor}} & \textbf{Optimizer} & \textbf{\textit{k} steps (\textit{H})} & \textbf{\makecell{\# Threads\\ (\textit{H}/\textit{L})}} & \textbf{\makecell{Training Time\\(\textit{H}/\textit{L})}}\\

\midrule
MARINE & Easy   & 0.8/0.1e7 & 5e-4  & 0.01 & 0.9 & Adam & 1 & 16/10 & 2.61/0.15 hrs \\
~   & Medium & 1.0/0.5e7 & 5e-4  & 0.01 & 0.9 & Adam & 3 & 16/10& 3.97/0.89 hrs \\
~   & Hard   & 3.4/0.6e7 & 5e-4  & 0.01 & 0.9 & Adam & 5 & 16/10& 11.04/1.06 hrs \\
\midrule
FC  & Easy   & 0.25/0.15e7 & 5e-4  & 0.01 & 0.9 & Adam & 1 & 16/10 & 0.81/0.37 hrs \\
~   & Medium & 0.3/0.5e7 & 5e-4  & 0.01 & 0.9 & Adam & 3 & 16/10 & 1.81/1.44 hrs \\
~   & Hard   & 0.5/1.3e7 & 5e-4  & 0.01 & 0.9 & Adam & 5 & 16/10 & 4.24/2.22 hrs \\
\bottomrule
\end{tabular}
}
\label{table:teacher_exp_details}
\vspace{-5pt}
\end{table*}
\subsection{MARINE}
\begin{figure}[t]
\centering
\includegraphics[width=0.46\textwidth]{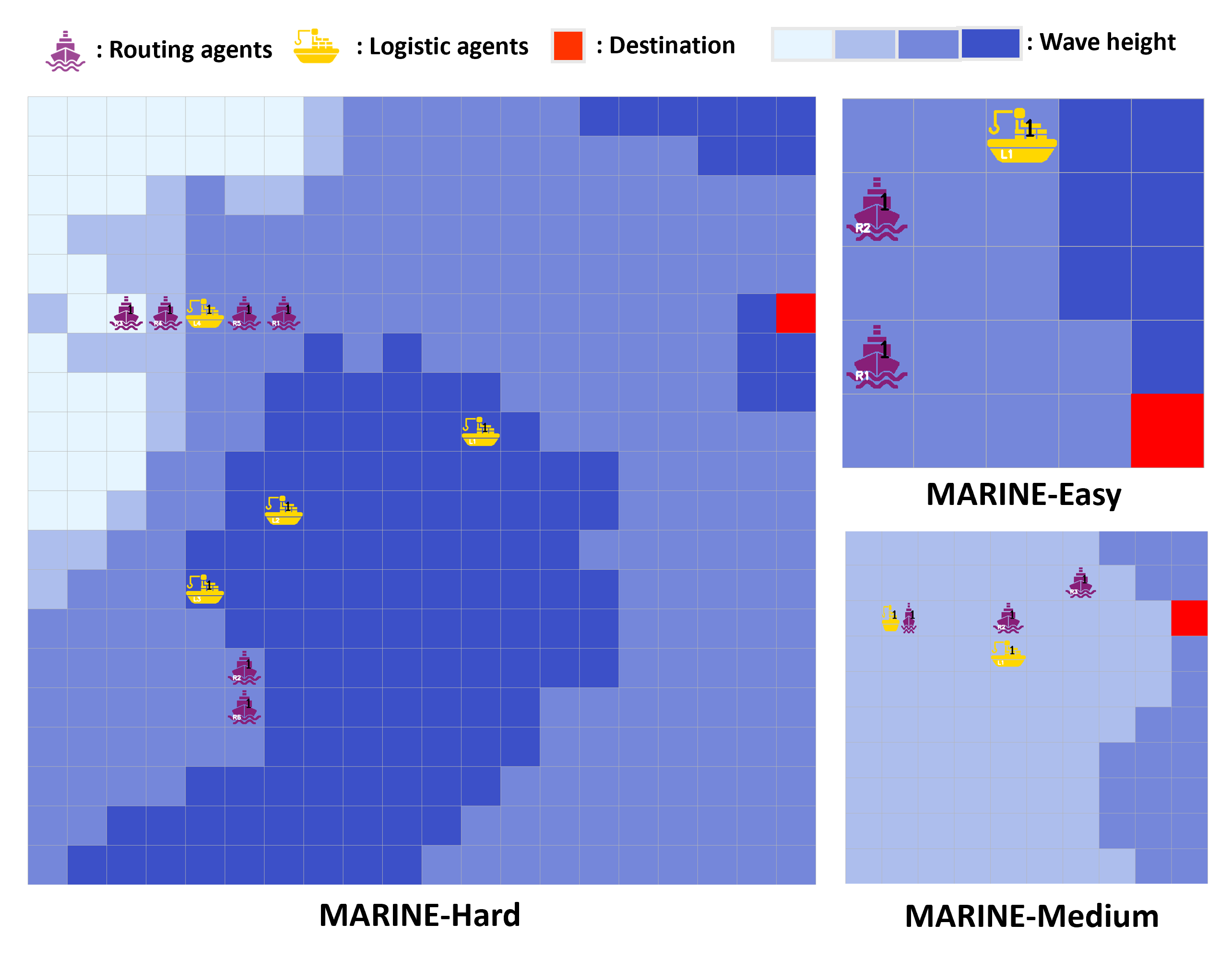}
\vspace{-10pt}
\caption{Illustrations of MARINE at varying difficulty levels (Easy, Medium, Hard).}
\label{fig:marine}
\end{figure}

A maritime logistics environment where routing agents traverse dynamic ocean conditions, while logistic agents provide mid-sea refueling support \rev{(See Fig.~\ref{fig:marine})}. The environment uses WaveWatch III forecast data, introducing weather-induced uncertainty into planning and coordination. Routing agents must reach their destinations before running out of fuel, as fuel depletion results in episode termination. 

Both routing and logistic agents operate under $n \times n$ partial observability, allowing them to sense nearby wave heights and agent positions. They share an action space comprising five primitive motions: $\texttt{up}$, $\texttt{down}$, $\texttt{left}$, $\texttt{right}$, and $\texttt{stay}$. Each movement, except for stay, consumes one unit of fuel. Refueling occurs when a routing agent and a logistic agent occupy the same grid cell. The efficiency of this refueling is determined by a four-level quantized wave height: no refueling takes place at the lowest level, while refueling up to 50\% of capacity is possible at the highest level.

The reward structure is designed to encourage coordination and discourage inactivity. Both types of agents incur small penalties at every time step, with a larger penalty applied if routing agents run out of fuel. Routing agents earn rewards for successfully reaching their destinations and are penalized based on their proximity to the goal. Logistic agents receive rewards proportional to the amount of fuel they transfer. Additional environment specifications are provided in Table~\ref{table:env_details}.

\subsection{FireCommander (FC)}
\begin{figure}[t]
\centering
\includegraphics[width=0.46\textwidth]{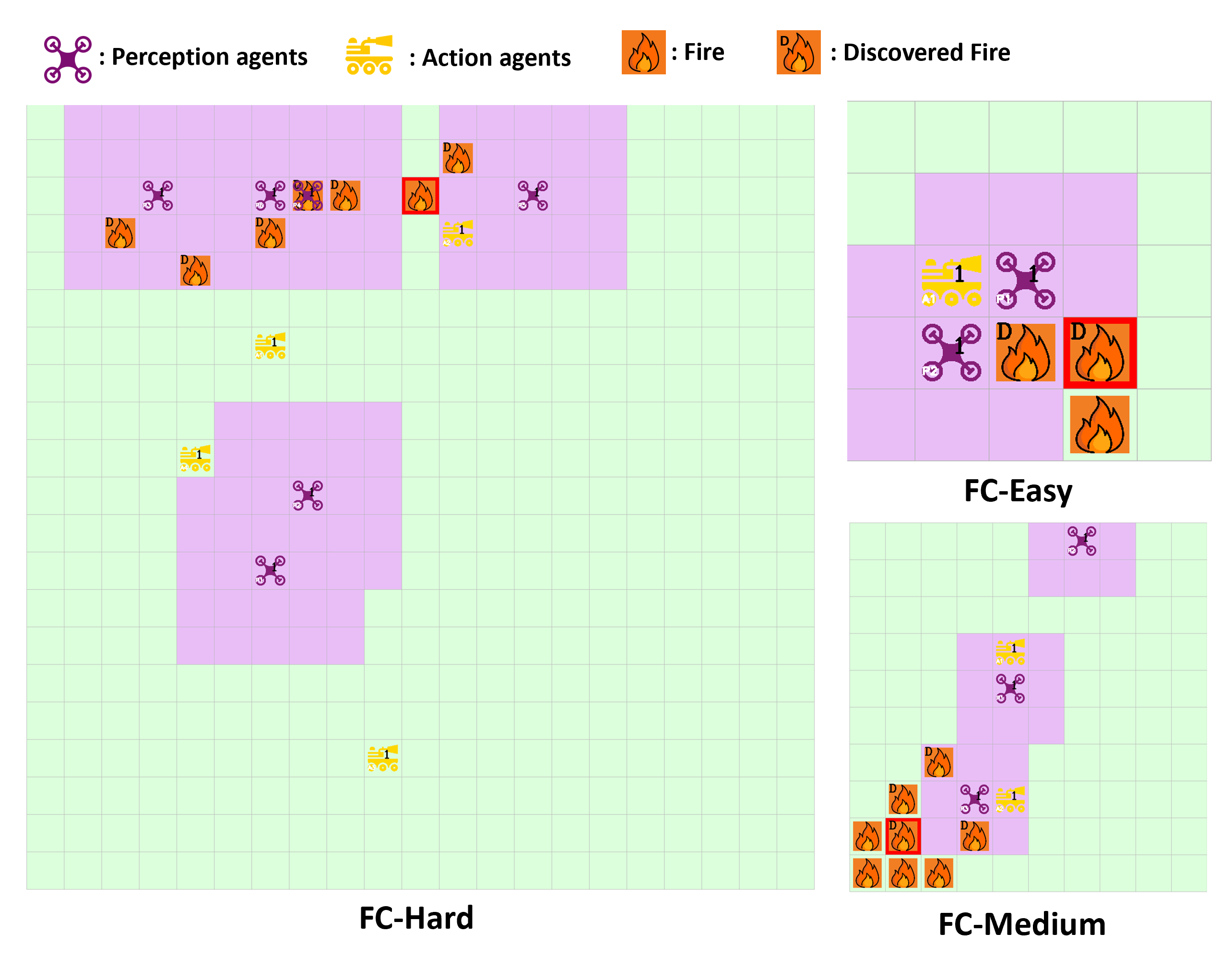}
\vspace{-10pt}
\caption{Illustrations of FireCommander (FC) at varying difficulty levels (Easy, Medium, Hard).}
\label{fig:firecommander}
\end{figure}

A grid-based wildfire environment inspired by the FARSITE model, where action agents can extinguish fires but lack perception capabilities, while perception agents monitor fire spread but cannot put out fires (\rev{See Fig.~\ref{fig:firecommander}}). The environment involves collaboration under partial observability and dynamic fire propagation, influenced by wind. Action agents can only extinguish fires that have been discovered by perception agents.

Perception agents operate under $n \times n$ partial observability, allowing them to sense the positions of nearby fires and agents, and share an action space consisting of five primitive motions: $\texttt{up}$, $\texttt{down}$, $\texttt{left}$, $\texttt{right}$, and $\texttt{stay}$. Action agents, \rev{which lack vision but have access to their global locations}, share these five primitive motions and also have a special action for $\texttt{putting out fires}$. \rev{In this work, to accelerate the monitoring of fire propagation, the entropy of fire probability information is newly introduced.}

The reward structure encourages cooperation and penalizes inactivity. Both perception and action agents receive small penalties proportional to the number of active fires. Action agents are rewarded for successfully extinguishing fires. Additional environment specifications are provided in Table~\ref{table:env_details}.

\section{Design Details of Teacher \& Student}
\label{Appendix:teacher and student models}

\subsection{Teacher Model: HiT-MAC}
\subsubsection{Formulate Hierarchical Structure}
\begin{figure*}[t]
\centering
\includegraphics[width=0.98\textwidth]{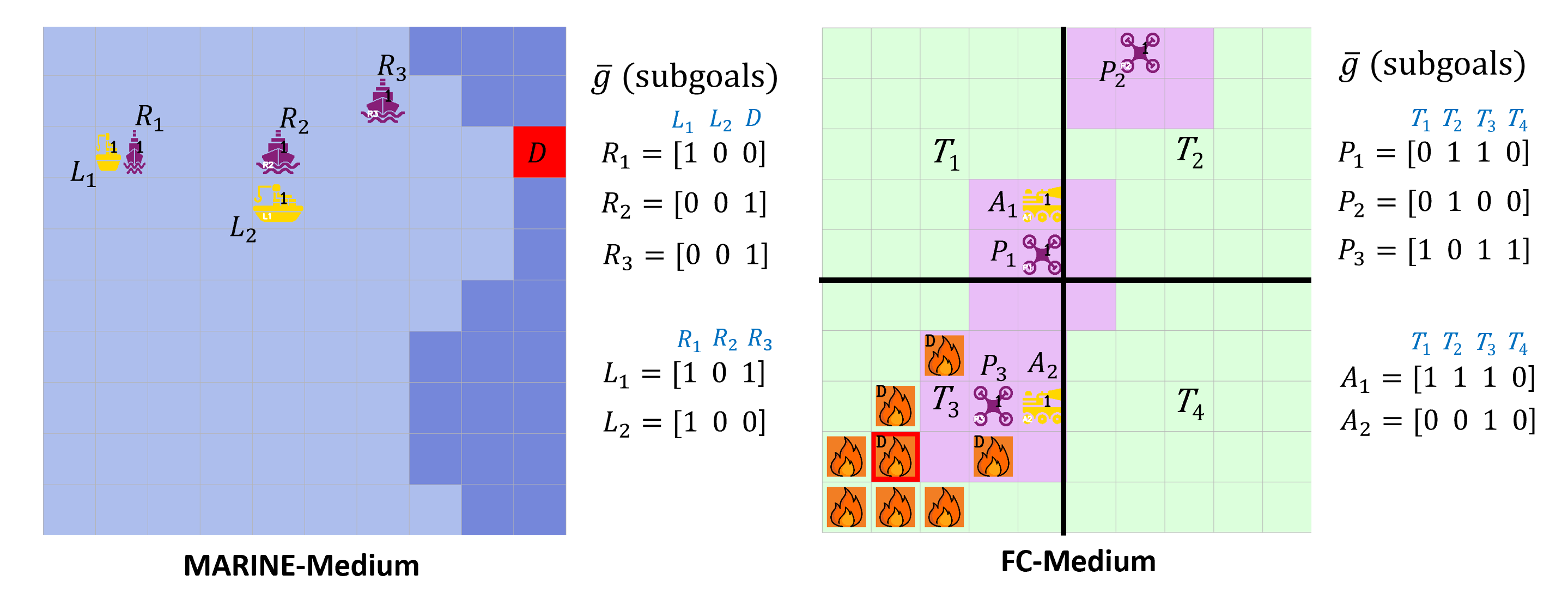}
\vspace{-15pt}
\caption{Examples of hierarchical structures in MARINE and FC (medium scenarios). In MARINE, \textit{R}, \textit{L}, and \textit{D} denote routing agents, logistic agents, and destinations, respectively. In FC, \textit{P}, \textit{A}, and \textit{T} represent perception agents, action agents, and subareas.}
\label{fig:hiearchy}
\vspace{-5pt}
\end{figure*}

We illustrate the formulation of hierarchical structures in MARINE and FC (See Fig.~\ref{fig:hiearchy}). In both domains, we handle heterogeneous observations and subgoals by employing class-specific encoders. To clearly distinguish subgoals assigned to different agent types, we use the agent's type character followed by a number, such as $R_1$ or $P_1$, instead of generic notation like $g_i$.

First, in the MARINE domain, routing agents aim to reach logistic agents or their destinations, while logistic agents focus solely on supporting routing agents. Routing agents choose behaviors based on their subgoals, such as heading directly to a destination when fuel levels are sufficient (e.g., $R_2=(0,0,1)$), or seeking refueling from logistic agents when fuel is low (e.g., $R_1=(1, 0, 0)$). Meanwhile, logistic agents may prioritize refueling low-fuel routing agents (e.g., $L_2 = (1, 0, 0)$) or supporting multiple routing agents at once (e.g., $L_1 = (1, 0, 1)$). All agent types share the same agent-target observation structure, which includes indices for both the agent and the target, fuel levels of the agent or target, the distance to the target, and local wave height information.

Next, in the FC domain, the environment is divided into multiple subareas that serve as shared targets for both perception agents and action agents—utilized for monitoring and fire-extinguishing tasks, respectively. Agents can cover multiple subareas broadly (e.g., $P_3 = (1, 0, 1, 1)$ or $A_1 = (1, 1, 1, 0)$), or focus narrowly on specific regions (e.g., $P_2 = (0, 1, 0, 0)$ or $A_2 = (0, 0, 1, 0)$). Agent-target observations typically include the indices of the agent and its assigned target. In addition, perception agents receive the maximum entropy location from the fire probability distribution within the subarea, while action agents receive the nearest discovered fire location and the density of discovered fires in the subarea.

\subsubsection{Training Setup} 
\begin{figure*}[t]
\centering
\includegraphics[width=0.99\textwidth]{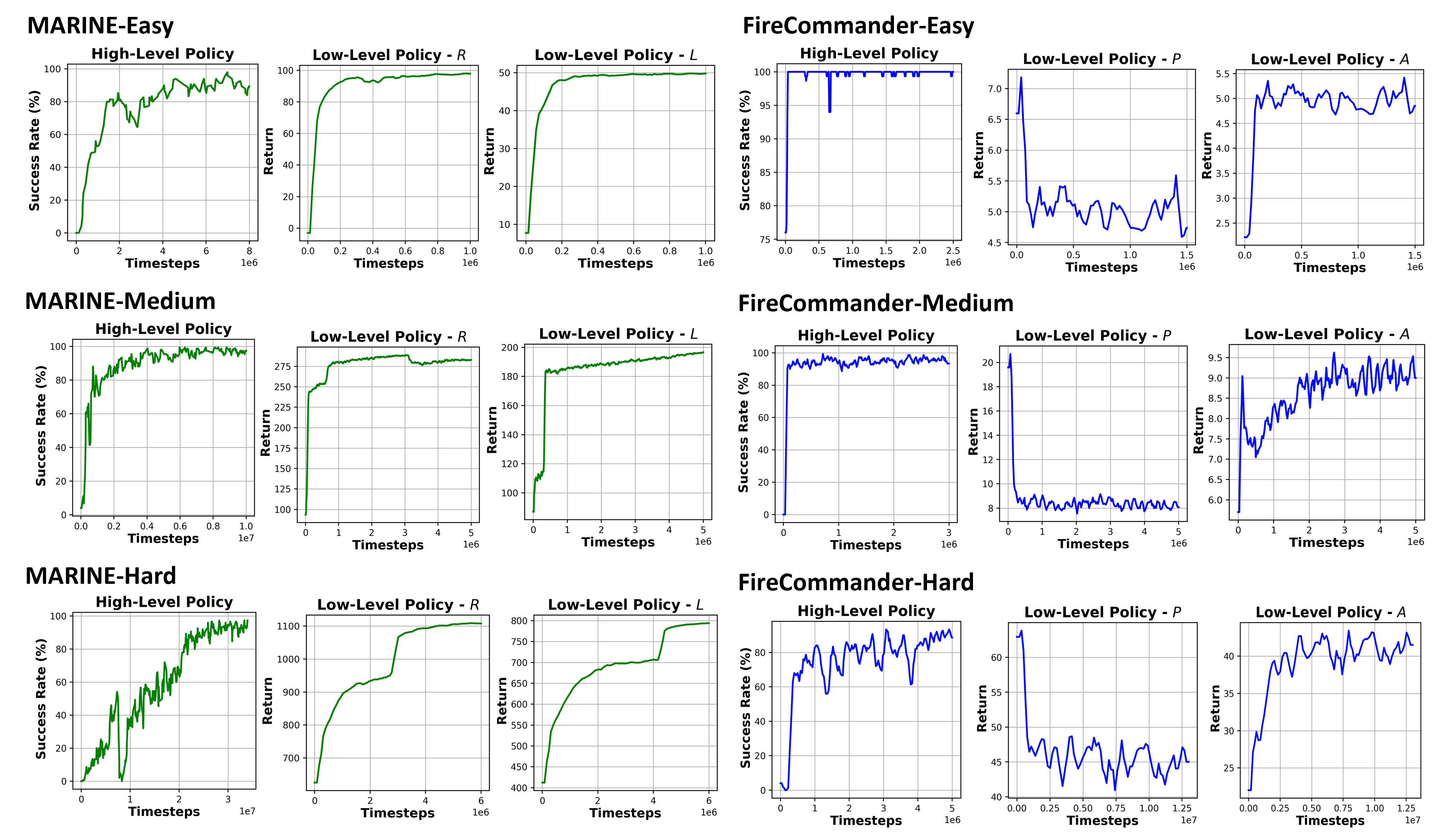}
\vspace{-12pt}
\caption{Learning curves of the teacher policy for MARINE and FireCommander. In MARINE, \textit{R} and \textit{L} denote the routing agents and logistic agents, respectively. In FireCommander, \textit{P} and \textit{A} represent the perception agents and action agents, respectively.}
\label{fig:teacher_curves}
\end{figure*}

For training, we adopt a two-stage strategy commonly used in hierarchical RL to improve stability and efficiency. First, low-level policies are trained independently to perform fundamental tasks effectively. Then, high-level policies are trained to coordinate these behaviors without being interrupted by immature low-level skills. Note that the high-level policy interacts with the low-level polices every $k$ steps to reduce computational overhead. During low-level training, we randomize agent positions and environmental factors, such as the number and location of fires, to encourage generalization. In this stage, agents are primarily trained to reach their assigned targets, with action agents in the FC domain receiving additional rewards for extinguishing fires. \rev{Note that the teacher’s training is based on the A3C framework~\cite{a3c}, with multiple threads employed to enhance computational efficiency. Please refer to Table~\ref{table:teacher_exp_details} for details of the teacher’s pre-training setup and Fig.~\ref{fig:teacher_curves} for the corresponding learning curves.}

\begin{figure*}[t]
\centering
\includegraphics[width=0.96\textwidth]
{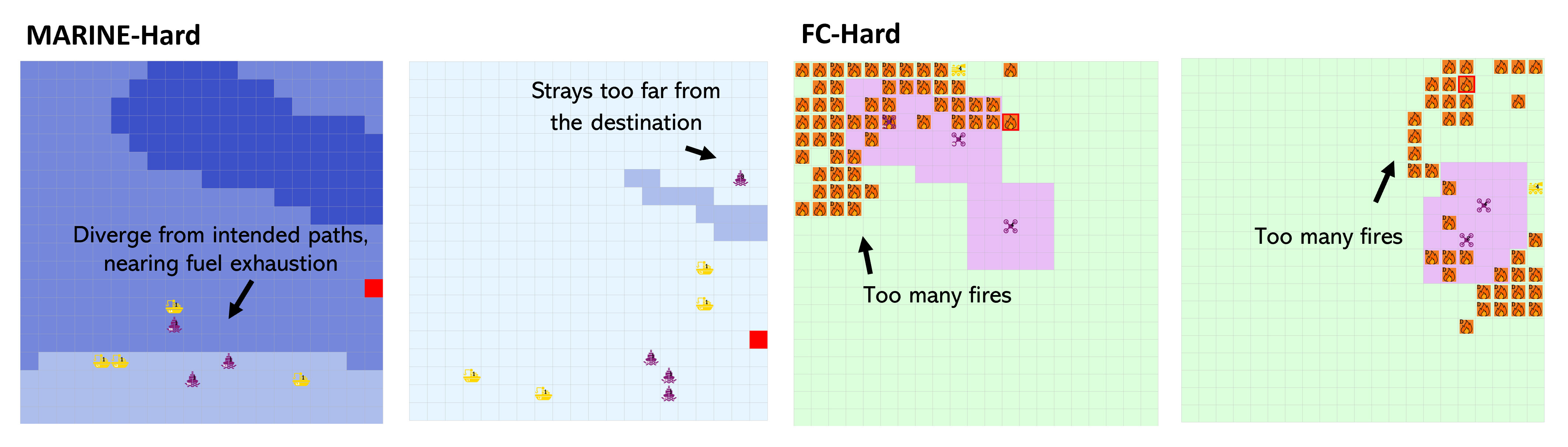}
\vspace{-12pt}
\caption{Examples where the teacher may provide suboptimal demonstrations due to student states deviating from the training distribution, particularly as the environment and team size scale up (e.g., MARINE-Hard, FC-Hard).}
\label{fig:hard_example}
\end{figure*}

\subsection{Student Model: HetNet}
We note that the latest version of HetNet~\cite{hetnet2} is used as the student model, as it supports tensorized observation structures that enable scalability across diverse team configurations and environment sizes. For example, if agents have a 3×3 field of view, nearby agent positions and environmental features such as wave height or fire positions can be captured in a structured tensor form. The preprocessing unit uses convolutional neural network (CNN) layers to process this spatial data. To support heterogeneous agents, the preprocessing unit, LSTM module, and decoder use class-specific weights that are shared among agents of the same type for efficiency. Furthermore, we incorporate two HetGAT layers, each with two-head multi-head attention, to enhance communication expressiveness and training stability.

\section{Backup: Student State Distribution When
Scaling Environments}
\label{Appendix:suboptimal data}




\rev{To compare the state distributions of the student and teacher in Fig.~\ref{fig:suboptimality}, we extract global context features from HiT-MAC’s self-attention module and project them into a common latent space via PCA. We then apply kernel density estimation (KDE)~\cite{kde} to estimate their distributions.} For both MARINE and FC, we sample 5,000, 10,000, and 15,000 states from teacher and student rollouts in the easy, medium, and hard scenarios, respectively.

{
\renewcommand{\arraystretch}{0.99}
\begin{table*}[t]
\caption{Training setup for MARINE. \textit{E}, \textit{M}, and \textit{H} corresponds to Easy, Medium, and Hard tasks.}
\vspace{-8pt}
\centering
\footnotesize
\resizebox{0.99\linewidth}{!}{
\begin{tabular}{l c c c c c c c c c c c c}
\toprule
\scriptsize \textbf{Method} & 
\scriptsize \makecell{\textbf{Timesteps}\\\textbf{(\textit{E, M, H}})} & 
\scriptsize \makecell{\textbf{Learning}\\\textbf{Rate}} & 
\scriptsize \makecell{\textbf{Discount}\\\textbf{Factor}} & 
\scriptsize \makecell{\textbf{Entropy}\\\textbf{Coefficient}} & 
\scriptsize \textbf{Optimizer} & 
\scriptsize \textbf{Batch Size} & 
\scriptsize \makecell{\textbf{\# Trajectories}\\\textbf{in Dataset}\\\textbf{(\textit{E, M, H}})} & 
\scriptsize \makecell{\textbf{\# Episodes for}\\\textbf{Expert Queries}\\\textbf{per Epoch}\\\textbf{(\textit{E, M, H}})} & 
\scriptsize \makecell{\textbf{\# Episodes for}\\\textbf{Teacher Training}\\\textbf{per Epoch}\\\textbf{(\textit{E, M, H}})} & 
\scriptsize \makecell{\textbf{\# Threads}\\\textbf{(\textit{E, M, H}})} &
\scriptsize \makecell{\textbf{Training Time}\\\textbf{(\textit{E, M, H}})} \\ 
\midrule
MAPPO      & 1.9/3.5/14e7 & 1e-4 & 0.97 & 0.01 & Adam       & 200 steps & -     & -     & -        & 10/20/20  & 11.57/8.69/58.34 hrs \\
TarMAC     & 1.9/3.5/14e7 & 1e-4 & 0.97 & 0.01 & RMSprop    & 200 steps & -     & -     & -          & 10/20/20  & 4.74/14.67/54.25 hrs \\
IC3Net     & 1.9/3.5/14e7 & 1e-4 & 0.97 & 0.01 & RMSprop    & 200 steps & -     & -     & -          & 10/20/20  & 6.46/13.21/69.80 hrs \\
CommNet    & 1.9/3.5/14e7 & 1e-4 & 0.97 & 0.01 & RMSprop    & 200 steps& -     & -     & -           & 10/20/20  & 4.37/13.91/67.90 hrs \\
HAPPO      & 1.9/3.5/14e7 & 1e-4 & 0.97 & 0.01 & Adam       & 200 steps& -     & -     & -           & 10/20/20  & 13.27/29.14/313.19 hrs \\
HetNet      & 1.9/3.5/14e7 & 1e-4 & 0.97 & 0.01 & Adam       & 200 steps& -     & -     & -         & 10/20/20   & 21.40/32.12/283.17 hrs \\
CTDS      & 5/8/20e7 & 1e-4 & 0.97 & 0.01 & Adam       & 10 episodes & -     & -     & -           & 10/20/20 & 31.12/27.20/305.12 hrs \\
IGM-DA      & 5/8/20e7 & 1e-4 & 0.97 & 0.01 & Adam       & 10 episodes& -     & -     & -       & 10/20/20    & 28.29/54.45/329.29 hrs \\
PTDE      & 5/8/20e7 & 1e-4 & 0.97 & 0.01 & Adam       & 10 episodes & -     & -     & -         & 10/20/20   & 63.86/31.42/230.08 hrs \\
HINT (Ours)       & 1/2/10e7 & 1e-4 & - & 0.01 & Adam       & 200 steps & 2000/3000/6000 & 20/100/200  & 10/150/600     & 10/20/20 & 11.46/12.18/133.90 hrs \\
\bottomrule
\end{tabular}
}
\label{table:ras_setup}
\end{table*}
}
{
\renewcommand{\arraystretch}{0.99}
\begin{table*}[t]
\caption{Training setup for FireCommander. \textit{E}, \textit{M}, and \textit{H} corresponds to Easy, Medium, and Hard tasks.}
\vspace{-8pt}
\centering
\footnotesize
\resizebox{0.99\linewidth}{!}{
\begin{tabular}{l c c c c c c c c c c c c}
\toprule
\scriptsize \textbf{Method} & 
\scriptsize \makecell{\textbf{Timesteps}\\\textbf{(\textit{E, M, H}})} & 
\scriptsize \makecell{\textbf{Learning}\\\textbf{Rate}} & 
\scriptsize \makecell{\textbf{Discount}\\\textbf{Factor}} & 
\scriptsize \makecell{\textbf{Entropy}\\\textbf{Coefficient}} & 
\scriptsize \textbf{Optimizer} & 
\scriptsize \textbf{Batch Size} & 
\scriptsize \makecell{\textbf{\# Trajectories}\\\textbf{in Dataset}\\\textbf{(\textit{E, M, H}})} & 
\scriptsize \makecell{\textbf{\# Episodes for}\\\textbf{Expert Queries}\\\textbf{per Epoch}\\\textbf{(\textit{E, M, H}})} & 
\scriptsize \makecell{\textbf{\# Episodes for}\\\textbf{Teacher Training}\\\textbf{per Epoch}\\\textbf{(\textit{E, M, H}})} & 
\scriptsize \makecell{\textbf{\# Threads}\\\textbf{(\textit{E, M, H}})} &
\scriptsize \makecell{\textbf{Training Time}\\\textbf{(\textit{E, M, H}})} \\ 
\midrule
MAPPO      & 1.4/2.8/8.8e7 & 1e-4 & 0.97 & 0.01 & Adam       & 200 steps & -     & -     & -        & 10/20/20  &  3.62/12.83/70.17 hrs \\
TarMAC     & 1.4/2.8/8.8e7 & 1e-4 & 0.97 & 0.01 & RMSprop    & 200 steps & -     & -     & -          & 10/20/20  & 5.58/13.76/42.98 hrs \\
IC3Net     & 1.4/2.8/8.8e7 & 1e-4 & 0.97 & 0.01 & RMSprop    & 200 steps & -     & -     & -          & 10/20/20  & 5.06/13.14/68.75 hrs \\
CommNet    & 1.4/2.8/8.8e7 & 1e-4 & 0.97 & 0.01 & RMSprop    & 200 steps& -     & -     & -           & 10/20/20  & 4.87/12.35/59.49 hrs \\
HAPPO      & 1.4/2.8/8.8e7 & 1e-4 & 0.97 & 0.01 & Adam       & 200 steps& -     & -     & -           & 10/20/20  & 3.98/14.3/172.31 hrs \\
HetNet      & 1.4/2.8/8.8e7 & 1e-4 & 0.97 & 0.01 & Adam       & 200 steps& -     & -     & -         & 10/20/20   & 11.80/33.07/178.78 hrs \\
CTDS      & 5/6/15e7 & 1e-4 & 0.97 & 0.01 & Adam       & 10 episodes & -     & -     & -           & 10/20/20 &  29.11/13.52/224.29 hrs\\
IGM-DA      & 5/6/15e7 & 1e-4 & 0.97 & 0.01 & Adam       & 10 episodes& -     & -     & -       & 10/20/20    &  24.00/18.43/248.90 hrs\\
PTDE      & 5/6/15e7 & 1e-4 & 0.97 & 0.01 & Adam       & 10 episodes & -     & -     & -         & 10/20/20   &  28.21/21.04/125.54 hrs\\
HINT (Ours)       & 1/2/7e7 & 1e-4 & - & 0.01 & Adam       & 200 steps & 1000/2000/4000 & 20/50/100  & 10/20/40     & 10/20/20 & 7.76/10.06/83.12 hrs \\
\bottomrule
\end{tabular}
}
\label{table:fc_setup}
\end{table*}
}

In Fig.~\ref{fig:hard_example}, we illustrate examples where student trajectories deviate from the training distribution, causing our hierarchical teacher to provide suboptimal demonstrations. First, in the MARINE-Hard environment, routing agents frequently deviate significantly from their intended paths toward destinations or refueling points and run out of fuel before meeting logistic agents, leaving them waiting until logistic agents arrive. The teacher effectively provides corrective guidance in smaller environments by directing logistic agents toward stranded routing agents. However, as the environment scales up, such scenarios become increasingly unfamiliar to the teacher because pre-trained low-level policies prevent students from significantly deviating from their paths during training. \rev{In turn, the teacher often defaults to idle behavior rather than actively coordinating refueling operations, and at times may even guide students to continue toward destinations without sufficient fuel.}

Similarly, in the FC-Hard scenario, student policies tend to underperform relative to the teacher policy, resulting in a substantial increase in the number of fires, situations rarely encountered by the teacher during training. In smaller environments, fires typically do not spread as widely, making such conditions less common. Since the teacher policies are generally not exposed to such escalating conditions, they struggle to respond effectively. This lack of exposure results in inefficient coordination among action agents, causing them to cluster in specific areas rather than distributing effectively across multiple subareas.

\section{Training Setup}
\label{Appendix:training steup}
In this section, we describe the experimental setup used for our benchmark evaluations and provide several additional results. Our method was trained on multiple machines, including systems equipped with an AMD EPYC 7513 CPU and NVIDIA A40 GPU, an Intel i9-13980HX CPU with an RTX 4070 Laptop GPU, \rev{an Intel i9-13900 CPU with an RTX 3070 GPU, an Intel Ultra 9 285K CPU with an RTX 5080 GPU}, and an Intel i7-12650H CPU with an RTX 3060 Laptop GPU. To ensure reproducibility, all experiments were conducted using identical conda environments with consistent library versions across all systems. For evaluation, results were \rev{tested on 50 episodes} and averaged over three different random seeds. Training setups across environment configurations are provided in Tables~\ref{table:ras_setup} and \ref{table:fc_setup}. \rev{Please note that CTDE baselines and HINT use 200 samples as the batch size for one gradient update, whereas KD baselines treat episodes as the batch unit, using 10 episodes per batch. In MARINE-Easy, KD baselines have an average of 15 steps per episode, while in all other settings, they experience at least 20 steps per episode on average. Thus, compared to the 200 timesteps used by HINT, KD baselines are trained with a larger number of samples.}


\section{Backup: Benchmark Test}
\label{app:full_benchmark}
\begin{figure*}[t]
  \centering
  \includegraphics[width=0.96\textwidth]{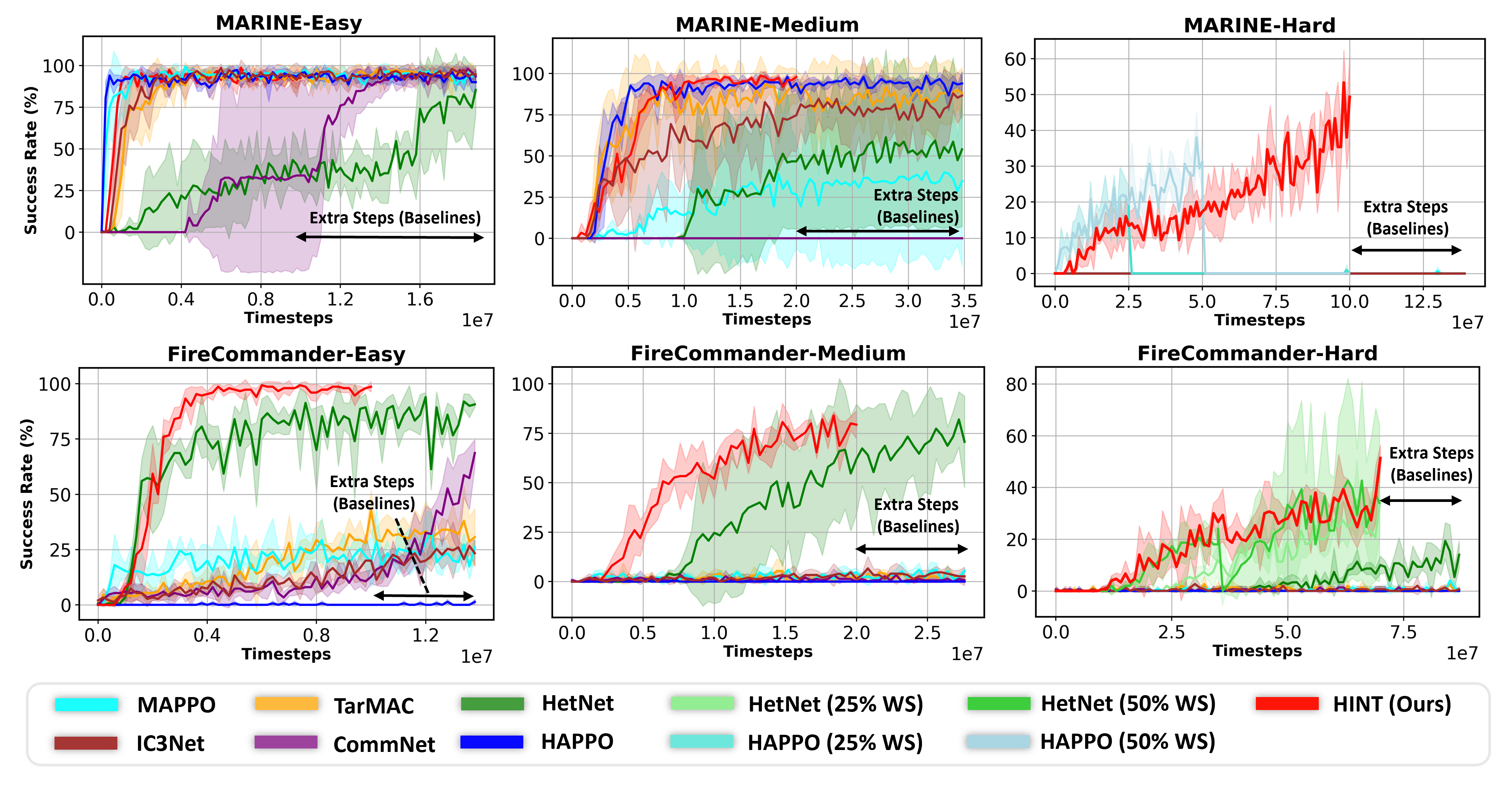}
  \vspace{-10pt}
  \caption{\rev{Learning curves for MARINE and FireCommander in the online CTDE benchmark, reported as the mean ($\pm$ standard deviation) across three random seeds and three difficulty settings. HINT consistently outperforms all baselines in both domains.}}
  \label{fig:fullonline_CTDE}
\end{figure*}

In this section, we present the full benchmark results. Tables ~\ref{table:full_bench_marine} and \ref{table:full_bench_fc} report benchmark tests for both CTDE and KD on MARINE and FireCommander (FC). Fig.~\ref{fig:fullonline_CTDE} illustrates the learning curves for MARINE and FireCommander in the online CTDE benchmark.
\begin{table*}[t]
\centering
\caption{Full quantitative results for FireCommander (FC) in the online CTDE and KD benchmark, reported as the mean ($\pm$ standard deviation) across three random seeds. \textit{WS} denotes warm starting RL with our teacher policy.}
\vspace{-7pt}
\resizebox{0.75\linewidth}{!}{
\begin{tabular}{lcccccc}
\toprule
\multirow{2}{*}{\textbf{Method}} & 
\multicolumn{2}{c}{\textbf{FC-Easy (5x5, 3 agents)}} & 
\multicolumn{2}{c}{\textbf{FC-Medium (10x10, 5 agents)}} & 
\multicolumn{2}{c}{\textbf{FC-Hard (21x21, 10 agents)}} \\
\cmidrule(lr){2-3}\cmidrule(lr){4-5}\cmidrule(lr){6-7}
 & Success Rate (\%)~$\uparrow$ & Steps Taken~$\downarrow$ 
 & Success Rate (\%)~$\uparrow$ & Steps Taken~$\downarrow$
 & Success Rate (\%)~$\uparrow$ & Steps Taken~$\downarrow$ \\
\midrule
MAPPO   & 32.67 $\pm$ 9.45 & 42.47 $\pm$ 1.90 & 6.00 $\pm$ 3.46 & 96.33 $\pm$ 1.19 & 4.00 $\pm$ 0.00 & 202.89 $\pm$ 0.86 \\
TarMAC   & 42.67 $\pm$ 9.02 & 42.20 $\pm$ 0.11 & 5.33 $\pm$ 5.03 & 96.89 $\pm$ 3.97 & 2.67 $\pm$ 3.06 & 207.51 $\pm$ 2.48 \\
IC3Net   & 26.67 $\pm$ 3.06 & 43.03 $\pm$ 0.46 & 6.67 $\pm$ 6.11 & 97.07 $\pm$ 2.79 & 2.67 $\pm$ 3.06 & 207.08 $\pm$ 2.68 \\
CommNet & 68.67 $\pm$ 6.11 & 34.44 $\pm$ 1.93 & 3.33 $\pm$ 3.06 & 97.81 $\pm$ 1.92 & 1.33 $\pm$ 2.31 & 207.96 $\pm$ 1.77 \\
HAPPO     & 1.33 $\pm$ 1.15 & 49.36 $\pm$ 0.55 & 0.67 $\pm$ 1.15 & 99.35 $\pm$ 1.13 & 0.67 $\pm$ 1.15 & 208.61 $\pm$ 2.40 \\
HetNet   & 94.00 $\pm$ 2.00 & 22.06 $\pm$ 3.02 & 82.00 $\pm$ 14.00 & 49.43 $\pm$ 16.00 & 19.33 $\pm$ 7.02 & 177.93 $\pm$ 11.91 \\
\hspace{0.1em}$\rightarrow$ 25\% \textit{WS}     & - & - & - & - & 35.33 $\pm$ 15.28 & 152.35 $\pm$ 18.78 \\
\hspace{0.1em}$\rightarrow$ 50\% \textit{WS}     & - & - & - & - & 42.67 $\pm$ 38.28 & 139.01 $\pm$ 62.69 \\
CTDS   & 18.89 $\pm$ 7.95 & 45.13 $\pm$ 3.04 & 1.67 $\pm$ 2.89 & 98.49 $\pm$ 2.61 & 0.83 $\pm$ 1.44 & 208.38 $\pm$ 2.80 \\
IGM-DA   & 16.22 $\pm$ 6.74 & 44.77 $\pm$ 2.13 & 2.78 $\pm$ 1.27 & 98.01 $\pm$ 2.09 & 1.11 $\pm$ 1.27 & 207.80 $\pm$ 2.59 \\
PTDE   & 24.00 $\pm$ 20.30 & 44.17 $\pm$ 4.51 & 3.61 $\pm$ 3.37 & 97.56 $\pm$ 3.22 & 2.71 $\pm$ 2.01 & 206.41 $\pm$ 4.25 \\
\textbf{HINT (Ours)} & \textbf{99.33 $\pm$ 1.15} & \textbf{15.69 $\pm$ 1.15} & \textbf{84.00 $\pm$ 2.00} & \textbf{47.95 $\pm$ 3.73} & \textbf{51.33 $\pm$ 5.03} & \textbf{136.40 $\pm$ 3.86} \\
\bottomrule
\end{tabular}
}
\label{table:full_bench_fc}
\end{table*}

\begin{table*}[t]
\centering
\caption{Full quantitative results for MARINE in the online CTDE and KD benchmark, reported as the mean ($\pm$ standard deviation) across three random seeds. \textit{WS} denotes warm starting RL with our teacher policy.}
\vspace{-7pt}
\resizebox{0.75\linewidth}{!}{
\begin{tabular}{lcccccc}
\toprule
\multirow{2}{*}{\textbf{Method}} & 
\multicolumn{2}{c}{\textbf{MARINE-Easy (5x5, 3 agents)}} & 
\multicolumn{2}{c}{\textbf{MARINE-Medium (10x10, 5 agents)}} & 
\multicolumn{2}{c}{\textbf{MARINE-Hard (20x20, 10 agents)}} \\
\cmidrule(lr){2-3}\cmidrule(lr){4-5}\cmidrule(lr){6-7}
 & Success Rate (\%)~$\uparrow$ & Steps Taken~$\downarrow$ 
 & Success Rate (\%)~$\uparrow$ & Steps Taken~$\downarrow$
 & Success Rate (\%)~$\uparrow$ & Steps Taken~$\downarrow$ \\
\midrule
MAPPO   & \textbf{99.33 $\pm$ 1.15} & 9.99 $\pm$ 0.38 & 40.67 $\pm$ 51.39 & 69.23 $\pm$ 37.31 & 0.67 $\pm$ 1.15 & 198.88 $\pm$ 1.94 \\
TarMAC   & 97.33 $\pm$ 3.06 & 9.43 $\pm$ 2.40 & 94.00 $\pm$ 10.39 & \textbf{25.84 $\pm$ 14.77} & 0.00 $\pm$ 0.00 & 200.0 $\pm$ 0.00 \\
IC3Net   & 98.67 $\pm$ 1.15 & \textbf{9.10 $\pm$ 2.39} & 88.00 $\pm$ 10.58 & 34.97 $\pm$ 17.99 & 0.00 $\pm$ 0.00 & 200.0 $\pm$ 0.00 \\
CommNet & 98.00 $\pm$ 2.00 & 11.09 $\pm$ 1.88 & 0.00 $\pm$ 0.00 & 100.00 $\pm$ 0.00 & 0.00 $\pm$ 0.00 & 200.0 $\pm$ 0.00 \\
HAPPO     & 97.33 $\pm$ 3.06 & 10.82 $\pm$ 1.04 & 98.67 $\pm$ 2.31 & 27.81 $\pm$ 0.46 & 0.00 $\pm$ 0.00 & 200.0 $\pm$ 0.00 \\
\hspace{0.1em}$\rightarrow$ 25\% \textit{WS}     & - & - & - & - & 24.00 $\pm$ 3.46 & 169.94 $\pm$ 9.25 \\
\hspace{0.1em}$\rightarrow$ 50\% \textit{WS}     & - & - & - & - & 38.00 $\pm$ 6.93 & 154.73 $\pm$ 8.30 \\
HetNet   & 85.33 $\pm$ 14.19 & 17.66 $\pm$ 5.02 & 61.33 $\pm$ 53.27 & 60.33 $\pm$ 34.64 & 0.00 $\pm$ 0.00 & 200.0 $\pm$ 0.00 \\
CTDS   & 76.00 $\pm$ 13.38 & 22.24 $\pm$ 1.69 & 23.06 $\pm$ 25.23 & 85.90 $\pm$ 11.16 & 7.22 $\pm$ 11.10 & 191.54 $\pm$ 13.06 \\
IGM-DA   & 93.33 $\pm$ 3.06 & 13.79 $\pm$ 1.42 & 80.00 $\pm$ 11.46 & 42.95 $\pm$ 7.40 & 14.17 $\pm$ 16.65 & 181.36 $\pm$ 21.99 \\
PTDE   & 98.22 $\pm$ 0.38 & 9.69 $\pm$ 1.47 & 96.11 $\pm$ 1.73 & 32.68 $\pm$ 1.01 & 33.33 $\pm$ 57.74 & 147.05 $\pm$ 91.71 \\
\textbf{HINT (Ours)} & 98.67 $\pm$ 1.15 & 11.39 $\pm$ 1.21 & \textbf{98.67 $\pm$ 1.15} & 26.49 $\pm$ 4.92 & \textbf{53.33 $\pm$ 9.02} & \textbf{143.28 $\pm$ 4.47} \\
\bottomrule
\end{tabular}
}
\label{table:full_bench_marine}
\end{table*}

\section{Back Up: Ablation Study}

\label{Appendix:ablation study}
\begin{figure*}[h]
  \centering
  \includegraphics[width=0.99\textwidth]{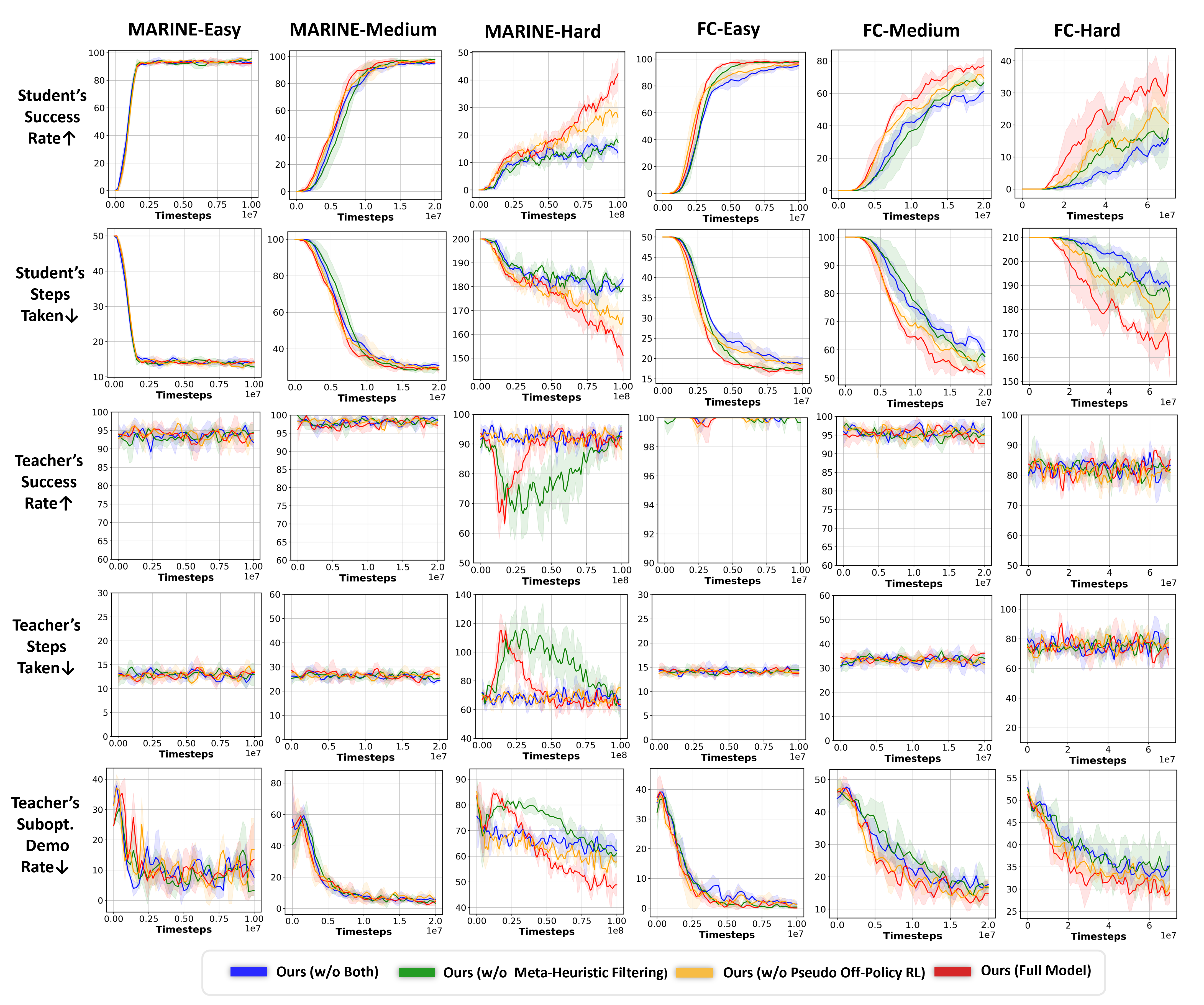}
  \caption{Full qualitative ablation results on MARINE and FireCommander (FC). Each column represents a domain-difficulty setting. Each row shows one evaluation metric over training timesteps (x-axis). Curves are averaged over three seeds, and shaded areas denote standard deviation (moving window length=5).}

  \label{fig:full_ablation}
  \vspace{-5pt}
\end{figure*}

\rev{As for the effect of key modules in HINT, we present full ablation results in Table~\ref{tab:full_ablation} and Fig.~\ref{fig:full_ablation}. Table~\ref{tab:full_ablation} reports quantitative results for MARINE and FireCommander across HINT and its three variants, while Fig.~\ref{fig:full_ablation} illustrates the training progress of the student and teacher across all environment settings. The results show that the adaptive teacher effectively assists the student’s training in more challenging domains, such as MARINE-Hard, FC-Medium, and FC-Hard.}

\begin{table*}[t]
\caption{Full quantitative results for MARINE and FireCommander (FC) in the ablation study, reported as the mean ($\pm$ standard deviation) across three random seeds.}
\label{tab:full_ablation}
\vspace{-5pt}
\centering
\footnotesize

\begin{tabular}{cc}
\centering
\resizebox{\linewidth}{!}{
\begin{tabular}{lcccccc}
\toprule
\multicolumn{1}{c}{\textbf{Method}} 
& \multicolumn{2}{c}{MARINE-Easy (5x5, 3 agents)} 
& \multicolumn{2}{c}{MARINE-Medium (10x10, 5 agents)} 
& \multicolumn{2}{c}{MARINE-Hard (20x20, 10 agents)} \\
\cmidrule(lr){2-3}\cmidrule(lr){4-5}\cmidrule(lr){6-7}
\textbf{} & Success Rate (\%)~\textuparrow & Steps Taken~\textdownarrow 
& Success Rate (\%)~\textuparrow & Steps Taken~\textdownarrow 
& Success Rate (\%)~\textuparrow & Steps Taken~\textdownarrow \\
\midrule
Ours (w/o Both) &97.33 ± 1.15&12.23 ± 0.91&96.67 ± 2.31&27.78 ± 3.76&22.00 ± 2.00&171.67 ± 2.42\\
Ours (w/o Performance-Based Filtering) &97.33 ± 1.15&11.95 ± 0.29&\textbf{99.33 ± 1.15}&26.81 ± 2.26&23.33 ± 13.32&168.49 ± 15.68\\
Ours (w/o Pseudo Off-Policy RL) &98.00 ± 0.00&\textbf{10.61 ± 1.92}&98.00 ± 2.00&\textbf{25.67 ± 0.75}&36.67 ± 10.26&156.71 ± 9.70\\
\textbf{Ours (Full Model)} &\textbf{98.67 ± 1.15}& 11.39 ± 1.21 &98.67 ± 1.15& 26.49 ± 4.92 &\textbf{53.33 ± 9.02} &\textbf{143.28 ± 4.47}\\
\bottomrule
\end{tabular}
}
\end{tabular}

\vspace{5pt} 

\begin{tabular}{cc}
\centering
\resizebox{\linewidth}{!}{
\begin{tabular}{lcccccc}
\toprule
\multicolumn{1}{c}{\textbf{Method}} 
& \multicolumn{2}{c}{FC-Easy (5x5, 3 agents)} 
& \multicolumn{2}{c}{FC-Medium (10x10, 5 agents)} 
& \multicolumn{2}{c}{FC-Hard (21x21, 10 agents)} \\
\cmidrule(lr){2-3}\cmidrule(lr){4-5}\cmidrule(lr){6-7}
\textbf{} & Success Rate (\%)~\textuparrow & Steps Taken~\textdownarrow 
& Success Rate (\%)~\textuparrow & Steps Taken~\textdownarrow 
& Success Rate (\%)~\textuparrow & Steps Taken~\textdownarrow \\
\midrule
Ours (w/o Both) &98.67 ± 1.15&17.87 ± 4.24&73.33 ± 7.57&53.09 ± 6.10&22.00 ± 13.11&178.41 ± 16.72\\
Ours (w/o Performance-Based Filtering) &\textbf{99.33 ± 1.15}&16.19 ± 1.48&77.33 ± 7.57&50.56 ± 8.06&28.00 ± 10.58&174.15 ± 12.53\\
Ours (w/o Pseudo Off-Policy RL) &\textbf{99.33 ± 1.15}&16.97 ± 2.29&76.67 ± 6.43&50.25 ± 7.11&28.67 ± 1.15&173.09 ± 1.61\\
\textbf{Ours (Full Model)} &\textbf{99.33 ± 1.15} &\textbf{15.69 ± 1.15} &\textbf{84.00 ± 2.00} &\textbf{47.95 ± 3.73} &\textbf{51.33 ± 5.03} &\textbf{136.40 ± 3.86}\\
\bottomrule
\end{tabular}
}
\end{tabular}

\end{table*}

\begin{table*}[t]
\renewcommand{\arraystretch}{0.8}
\centering
\caption{Performance and training time comparison on MARINE-Hard and FC-Hard benchmarks. For HINT, pre-training time is included in the total training time. PTDE’s training time is interpolated to match the CTDE baseline timesteps.}
\label{tab:benchamrk_time_comparision}
\vspace{-5pt}
\resizebox{0.98\linewidth}{!}{
\begin{tabular}{lcccccccc}
\toprule
\multicolumn{1}{c}{\textbf{Method}} 
& \multicolumn{4}{c}{MARINE-Hard} 
& \multicolumn{4}{c}{FireCommander-Hard} \\
\cmidrule(lr){2-5}\cmidrule(lr){6-9}
& Success Rate (\%)~\textuparrow & $\Delta$ (\%) & Training Time (hr)~\textdownarrow & $\Delta$ (\%) 
& Success Rate (\%)~\textuparrow & $\Delta$ (\%) & Training Time (hr)~\textdownarrow & $\Delta$ (\%) \\
\midrule
HAPPO   & 0.00 & -100\%  & 313.19 & +83\% & 0.67 & -99\% & 172.31 & +92\% \\
HetNet  & 0.00 & -100\% & 283.17& +65\% & 19.33 & -62\% & 178.78 & +100\% \\
PTDE  & 33.33 & -37\% & 161.06 & +10\% & 2.71 & -94\% & 73.65 & -18\% \\
\textbf{HINT} & \textbf{53.33} & --- & \textbf{146.00} & --- & \textbf{51.33} & --- & \textbf{89.58} & --- \\
\bottomrule
\end{tabular}
}
\end{table*}

\begin{table*}[t]

\centering
\caption{Effect of HINT modules on performance and training time in MARINE-Hard and FC-Hard tasks.}
\label{tab:ablation_time_comparision}
\resizebox{0.98\linewidth}{!}{
\begin{tabular}{lcccccccc}
\toprule
\multicolumn{1}{c}{\textbf{Method}} 
& \multicolumn{4}{c}{MARINE-Hard} 
& \multicolumn{4}{c}{FireCommander-Hard} \\
\cmidrule(lr){2-5}\cmidrule(lr){6-9}
& Success Rate (\%)~\textuparrow & $\Delta$ (\%) & Training Time (hr)~\textdownarrow & $\Delta$ (\%) 
& Success Rate (\%)~\textuparrow & $\Delta$ (\%) & Training Time (hr)~\textdownarrow & $\Delta$ (\%) \\
\midrule
HINT (w/ only communication; BC)   & 15.33 & -71\%  & 65.46 & -55\% & 28.67 & -44\% & 42.93 & -52\% \\
HINT (w/o Pseudo Off-Policy RL)  & 36.67 & -31\% & 149.43 & +2\% & 28.67 & -44\% & 76.02 & -15\% \\
\textbf{HINT (Full Model)} & \textbf{53.33} & --- & \textbf{146.00} & --- & \textbf{51.33} & --- & \textbf{89.58} & --- \\
\bottomrule
\end{tabular}
}
\end{table*}

\begin{figure*}[t]
  \centering
  \includegraphics[width=0.98\linewidth]{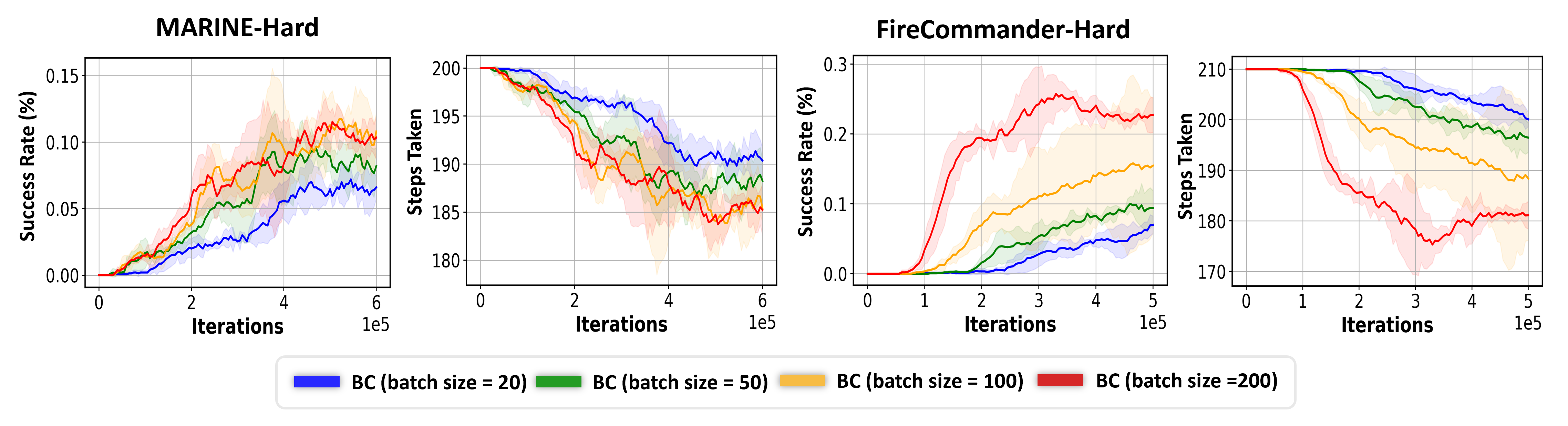}
  \vspace{-10pt}
  \caption{Behavior cloning results on MARINE-Hard and FireCommander-Hard across different batch sizes, reported as the mean (± standard
deviation). Each iteration corresponds to one update step using a batch of data.}
  \label{fig:batch_update}
  \vspace{-5pt}
\end{figure*}

We examine the effect of batch size on training stability using behavior cloning (BC) in Fig.~\ref{fig:batch_update}. BC uses the same student policy as HINT and the same teacher demonstrations. To isolate the impact of each update, we plot metrics over training iterations, with each iteration corresponding to one batch update. In both MARINE-Hard and FC-Hard, larger batch sizes lead to more stable and effective learning.
\section{Training Time Comparison}
\label{app:time_analysis}
\rev{As discussed in Sec.~\ref{sec:limit, broader}, HINT incorporates diverse modules such as pseudo-off policy RL, performance-based filtering, dataset aggregation, and HetGAT-based communication, which may lead to higher computational costs. Here, we compare HINT’s performance and training time with those of competitive baselines (e.g., HAPPO, HetNet, and PTDE) on the MARINE-Hard and FC-Hard tasks. Note that we include the pre-training time in HINT’s total training time. For PTDE, the training time is interpolated to match the CTDE baseline timesteps, as it was trained with significantly more timesteps than the CTDE baselines and HINT. Table~\ref{tab:benchamrk_time_comparision} shows that HINT achieves faster training time compared to HAPPO and HetNet while delivering significant performance improvements. Moreover, HINT demonstrates comparable training time to PTDE while achieving substantial gains in performance.}

\rev{To investigate the impact of each key module on HINT’s training time, we compare the training time of HINT with its variants, as shown in Table~\ref{tab:ablation_time_comparision}. Most variants executed the teacher policy through full episodes during dataset aggregation to track suboptimal demonstration rates (Fig.~\ref{fig:full_ablation}), which makes it difficult to isolate the exact cost of performance-based filtering. Alternatively, we compare the full model with two specific variants: one that includes only communication (e.g., behavior cloning, BC) and another that excludes pseudo off-policy RL. This comparison reveals both the overall effect of our interactive distillation modules (pseudo off-policy RL, performance-based filtering, and dataset aggregation) and the isolated effect of pseudo off-policy RL.}

\rev{Overall, the full model requires approximately twice the training time of BC, indicating that the interactive distillation modules contribute significantly to the training overhead. However, BC experiences a performance drop of 40–70\%, which justifies the use of interactive distillation despite its added cost. When comparing the full model to the version without pseudo off-policy RL, the training times are similar. This suggests that most of the overhead comes from dataset aggregation and performance-based filtering rather than from the pseudo off-policy RL itself. These findings highlight the need for a more sophisticated inference module to evaluate the quality of teacher demonstrations, rather than relying on resource-intensive performance-based filtering. Such enhancements could significantly lower computational costs while still achieving strong performance.}

\end{document}